\documentclass{article}

\usepackage{hyperref}

\usepackage{PRIMEarxiv}
\usepackage[utf8]{inputenc} 
\usepackage[T1]{fontenc}    
\usepackage{url}            
\usepackage{booktabs}       
\usepackage{amsfonts}       
\usepackage{nicefrac}       
\usepackage{fancyhdr}       
\usepackage{graphicx}       
\graphicspath{{media/}}     

\usepackage{xcolor}         

\usepackage{amsmath}
\usepackage{cleveref}
\usepackage{algorithmic}
\usepackage{algorithm}
\usepackage{tikz}
\usepackage{lscape}
\usepackage{arydshln}
\usepackage{pifont}
\usepackage{pgfplots}
\usepackage{pgfplotstable}
\pgfplotsset{compat=newest}
\usepackage{nicematrix}
\usepackage{rotating}
\usepackage{dsfont}
\usepackage{multirow}
\usepackage{array}
\usepackage{amssymb}
\usepackage[export]{adjustbox}

\usepackage{tabularx, ragged2e}
\newcolumntype{Y}{>{\centering\arraybackslash}X} 

\usepackage{afterpage}
\usepackage{setspace}
\usepackage{ntheorem}

\newtheorem{proposition}{Proposition}
\newtheorem{lemma}{Lemma}

\usepackage{pgfplots}
\usepackage{pgfkeys}
\usepgflibrary{shapes.geometric}

\theoremstyle{nonumberplain}
\theorembodyfont{\normalfont}
\theoremsymbol{\ensuremath\square}
\newtheorem{proof}{Proof:}

\definecolor{dodgerblue}{RGB}{30,144,255}
\definecolor{greenG}{RGB}{0,128,0}
\definecolor{color1}{RGB}{196, 164, 132} 
\definecolor{color2}{RGB}{30,144,255} 
\definecolor{color3}{RGB}{255, 16, 240} 

\usepackage{multirow}  
\usepackage{graphicx}  

\title{A unified framework of non-local parametric methods for image denoising}

\author{Sébastien Herbreteau, Charles Kervrann \\
Centre Inria de l'Université de Rennes, France\\
\texttt{\{sebastien.herbreteau, charles.kervrann\}@inria.fr} \\
}

\begin{document}

\maketitle

\begin{abstract}
We propose a unified view of  non-local methods
for single-image denoising, for which BM3D is the most popular representative, that operate
by gathering noisy patches together according to their similarities in order to
process them collaboratively. Our general estimation framework is based on the
minimization of the quadratic risk, which is approximated in two steps, and
adapts to photon and electronic noises. Relying on unbiased risk estimation (URE) for the
first step and on ``internal adaptation'', a concept borrowed from deep learning
theory, for the second, we show that our approach enables to reinterpret and
reconcile previous state-of-the-art non-local methods. Within
this framework, we propose a novel  denoiser called NL-Ridge that exploits linear
combinations of patches. While conceptually simpler, we show that NL-Ridge
can outperform well-established state-of-the-art single-image denoisers.
\end{abstract}

\section{Introduction} 

Over the years, a diverse range of strategies, tools, and theories have been proposed to solve the image denoising problem, at the crossroads of statistics, signal processing, optimization, and functional analysis. However, the recent immense influence on this field stems from the advancements in machine learning techniques, particularly deep neural networks. By framing denoising as a regression problem, the task essentially involves training a network to map the corrupted image to its source. This paradigm shift has revolutionized denoising and various other inverse problems in computer vision. Since then, numerous supervised neural networks have been introduced for image denoising \cite{dncnn, ffdnet, LIDIA, drunet, restormer, scunet, swinir, red30, tnrd, mwcnn, nlrn, n3net}, achieving state-of-the-art performances.

However, these supervised approaches, apart from their demanding optimization phase, face challenges due to their heightened sensitivity to the quality of the training data. These methods require a training dataset that not only contains a diverse range of examples but also must be abundant and representative of various image scenarios. Otherwise, subpar or entirely inaccurate outcomes may result. Consequently, their application becomes impractical in certain scenarios, particularly when noise-free images are unavailable (although training on datasets  composed of noisy/noisy image pairs was studied in  \cite{noise2noise, noisier2noise, R2R, N2V, laine, N2S}). In an attempt to overcome this hurdle, single-image learning—an approach in which only the input noisy image is utilized for training—was explored with deep neural networks \cite{S2S, DIP, N2S, ZS-N2N, rethinking, N2F} as an alternative strategy. However, their performance is still limited compared to their conventional counterparts \cite{BM3D, nlbayes, WNNM, NCSR, SAIST, ksvd, PEWA, OWF, TWSC, EPLL_unsupervised}. Within dataset-free image denoising, BM3D \cite{BM3D} continues to serve as a benchmark method and maintains competitiveness even though it was introduced approximately fifteen years ago. Relying on the non-local strategy, the mechanism of BM3D involves collaboratively processing groups of similar noisy patches distributed throughout the image. Subsequently, numerous methods rooted in patch grouping have emerged, such as \cite{nlbayes, WNNM, SAIST, NCSR, TWSC}.

In this paper, we present a unified view, based on quadratic risk minimization, to reinterpret and reconcile two-step non-local methods \cite{nlbayes, BM3D} from families of parameterized functions. We focused on BM3D \cite{BM3D} and NL-Bayes \cite{nlbayes} as they are considered as the best performing and fastest unsupervised methods in image denoising so far. In our estimation framework, no prior model for the distribution on patches is required. Second, derived from this framework, we propose a novel denoiser called NL-Ridge that exploits linear combinations of patches. We show that the resulting algorithm may outperform BM3D \cite{BM3D} and NL-Bayes \cite{nlbayes}, as well as several single-image deep-learning methods \cite{S2S, DIP, N2S, ZS-N2N, rethinking, N2F}, while being simpler conceptually.

The remainder of this paper is organized as follows. In
\Cref{nlridge_section2}, we construct NL-Ridge algorithm from the family of parameterized functions that processes patches by linear combinations, whether constrained or not. In
 \Cref{nlridge_section3},  we show that when considering two specific families of functions,  NL-Bayes \cite{nlbayes} and BM3D \cite{BM3D} can be fully reinterpreted within our statistical framework. Finally, in \Cref{nlridge_section4}, we demonstrate on artificially noisy but also real-noisy images that NL-Ridge compares favorably with its well-established counterparts \cite{BM3D,nlbayes}, despite relying only on linear combinations of patches.

 \begin{figure*}[t]
\centering
\includegraphics[width=0.9\columnwidth]{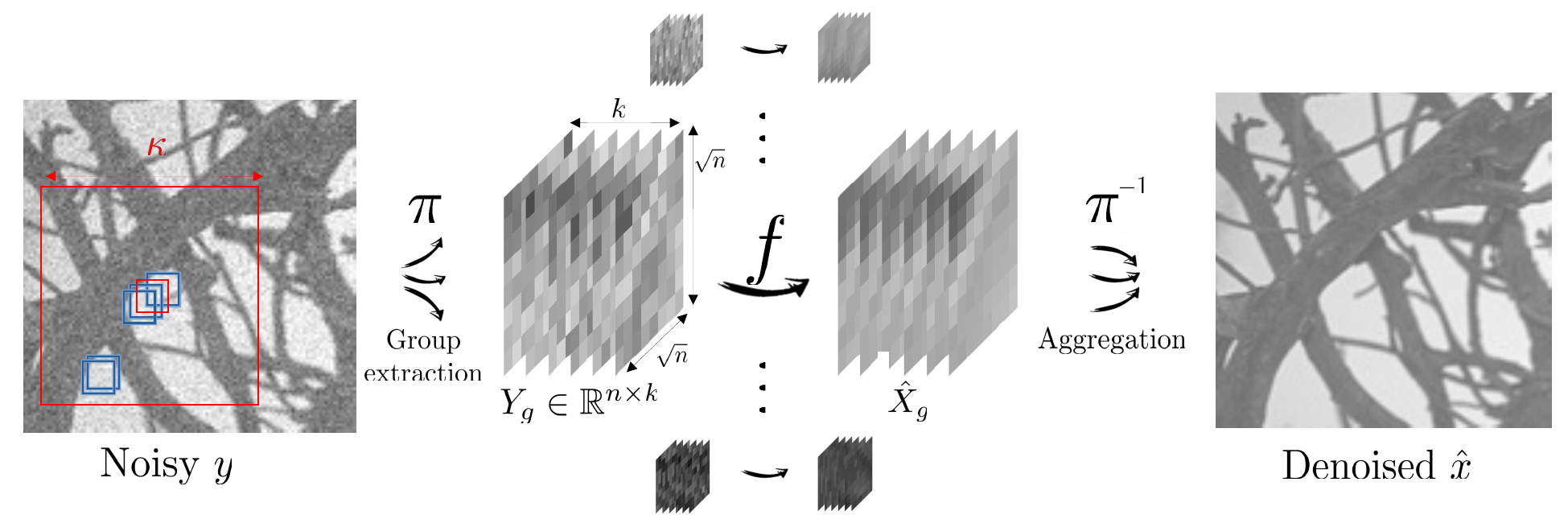}
\caption[Grouping technique for image denoising]{Illustration of the grouping technique for image denoising.}
\label{bm0}
\end{figure*}
\section{NL-Ridge for image denoising}
\label{nlridge_section2}

Popularized by BM3D \cite{BM3D}, the grouping technique, sometimes referred to as block-matching, has proven to be a key element in achieving state-of-the-art performances in single-image denoising \cite{BM3D, nlbayes, WNNM, SAIST, LSSC, NCSR, PLR, cuisine}. This technique consists in gathering noisy patches together according to their similarities in order to denoise them collaboratively. First, groups of $k$ similar noisy square patches $\sqrt{n} \times \sqrt{n}$ are extracted from the noisy image $y$. Specifically, for each overlapping patch $y_g$ taken as reference, the similarity (\textit{e.g.}, in the $\ell_2$ sense) with its surrounding overlapping patches (\textit{e.g.}, within a local window $\kappa \times \kappa$) is computed and the $k$-nearest neighbors, including the reference patch, are then selected to form a so-called similarity matrix $Y_g \in \mathbb{R}^{n \times k}$, where each column represents a flattened patch. Subsequently, all groups are processed in parallel by applying a local denoising function $f$ that produces an estimate for each noise-free similarity matrix: $\hat{X}_g = f(Y_g) \in \mathbb{R}^{n \times k}$. Finally, the denoised patches are repositioned to their initial locations in the image and aggregated, or reprojected \cite{Aggreg}, as pixels may have several estimates, to build the final estimated image $\hat{\mathcal{I}}$. Generally, arithmetic (sometimes weighed) averaging of all estimates for a same underlying pixel is used to that end.  \Cref{bm0} displays the whole process, summarized in \Cref{algo_nlridge_basic} for a given local denoising function $f$.

\begin{algorithm}
\caption{Non-local methods for image denoising}
\label{algo_nlridge_basic}
\begin{algorithmic}
\STATE{\textbf{Input:} Noisy image $y$, patch size $\sqrt{n}$, group size $k$}

\FOR{each $\sqrt{n} \times \sqrt{n}$ overlapping noisy patch in $y$}
\STATE{Extract its $k$ most similar patches to form similarity matrix $Y_g$}
\STATE{Perform collaborative denoising $\hat{X}_g = f(Y_g)$}
\ENDFOR

\STATE{Aggregate all the denoised patches contained in the groups $\hat{X}_g$ to form the  image $\hat{\mathcal{I}}$}

\RETURN{$\hat{\mathcal{I}}$}
\end{algorithmic}
\end{algorithm}

Within this framework, the choice of the local denoising function $f$ remains an open question. A majority of state-of-the-art methods leverage the inherent sparsity of natural images for the design of $f$ \cite{BM3D, LSSC, WNNM, SAIST, NCSR, PLR}. For example, BM3D \cite{BM3D} and LSSC \cite{LSSC} assume a locally sparse representation of the similarity matrices in a predetermined basis or dictionary (wavelets or DCT) while others rather adopt a low-rank approach \cite{WNNM, SAIST, NCSR, PLR}. As for NL-Bayes \cite{nlbayes}, a Bayesian framework is exploited at the patch level, in which $f$ produces a maximum a posteriori probability (MAP) estimate.

\subsection{Parametric linear patch combinations}

Focusing on functions that perform locally linear combinations of patches,  $f$ is chosen among the set of the following parameterized functions for each input similarity matrix $Y_g$: 
\begin{equation}
f_{\Theta} : Y \in \mathbb{R}^{n \times k} \mapsto Y \Theta
\label{local_denoiser0}
\end{equation}
\noindent where $\Theta \in \mathbb{R}^{k \times k}$.
Essentially, the $k$ columns of matrix $\Theta$ encode the weights of the $k$ different linear combinations aimed to be applied for patch group denoising. Note that parameters $\Theta$ may change from one similarity matrix $Y_g$ to another, so that as many different matrices $\Theta$ as there are similarity matrices $Y_g$ may be chosen. According to the constraints imposed on the combination weights, the search space for parameters $\Theta$ is restricted to subsets of $\mathbb{R}^{k \times k}$ as follows ($\mathbf{1}_k$ denotes the $k$-dimensional all-ones vector):


\begin{itemize}
    \item \textbf{linear combinations} of patches: $\Theta \in \mathbb{R}^{k \times k}$.
    \item \textbf{affine combinations} of patches: $\Theta \in \mathbb{R}^{k \times k}$ s.t.   $\Theta^\top \mathbf{1}_k = \mathbf{1}_k$.
    \item \textbf{conical combinations} of patches: $\Theta \in \mathbb{R}^{k \times k}$ s.t.  $\Theta \succeq 0$.
    \item \textbf{convex combinations} of patches:  $\Theta \in \mathbb{R}^{k \times k}$ s.t.  $\Theta^\top \mathbf{1}_k = \mathbf{1}_k$ and $\Theta \succeq 0$.
\end{itemize}

Aggregating similar patches  via a linear (in general convex) combination has already been exploited in the past \cite{nlmeans, PEWA, OWF, nlmeans_parameters}. However, the originality of our approach lies in the way of computing the weights $\Theta$, which significantly boosts performance.

\subsection{Parameter optimization}

In what follows, for the sake of notation simplicity, the index $g$ from $Y_g$ designating the group patch is removed. Thus, $Y \in \mathbb{R}^{n \times k}$ denotes any similarity matrix resulting of the corruption of $X$, its associated clean similarity matrix, by the underlying noise model (\textit{e.g.}, Gaussian noise, Poisson noise, \ldots). 

For each patch group $Y$, $f_\Theta$  is found by minimizing the local quadratic risk defined as:
\begin{equation}
\mathcal{R}_\Theta(X) = \mathbb{E}\|f_\Theta(Y) - X\|^2_F\,.
\label{riskDef}
\end{equation}
In other words, we look for the Minimum-Mean-Squared-Error (MMSE) estimator  $f_{\Theta^\ast}$ among the family of functions $(f_\Theta)_{\Theta \in \mathbb{R}^{k \times k}}$ defined in \eqref{local_denoiser0}. The optimal estimator $f_{\Theta^\ast}$ minimizes the risk, \textit{i.e.}
\begin{equation}
    \Theta^\ast = \arg \min_{\Theta} \;  \mathcal{R}_{\Theta}(X).
\label{eq2}
\end{equation}
Unfortunately, $\Theta^\ast$ requires the knowledge of $X$  which is unknown. The good news is that the risk $\mathcal{R}_{\Theta}(X)$  can be approximated through the following two-step algorithm:

\begin{enumerate}
    \item In Step 1, an approximation of $\Theta^\ast$ is computed for each group of similar patches, through the use of an unbiased risk estimate (URE) of $\mathcal{R}_{\Theta}(X)$. After reprojection \cite{Aggreg} of all denoised patches, a first denoised image $\hat{\mathcal{I}}^{(1)}$ is obtained.
    \item In Step 2, $\hat{\mathcal{I}}^{(1)}$ is improved with a second estimation of $\Theta^\ast$ which is found thanks to the technique of ``internal adaptation'' described in \cite{LIDIA} to eventually form $\hat{\mathcal{I}}^{(2)}$.
\end{enumerate}

\bigskip 
\noindent In what follows, we focus on three different types of noise:

\begin{itemize}
    \item Gaussian noise: $Y_{i,j} \sim \mathcal{N}(X_{i,j}, V_{i,j})$ with $V \in (\mathbb{R}_\ast^+)^{n \times k}$ indicating the variance per pixel, sometimes referred to as ``noisemap''. In particular, for homoscedastic Gaussian noise,  $V = \sigma^2 \mathbf{1}_n \mathbf{1}_k^\top$ where $\sigma > 0$ is the standard deviation, that is $Y_{i,j} \sim \mathcal{N}(X_{i,j}, \sigma^2)$,  
    \item Poisson noise: $Y_{i,j} \sim \mathcal{P}(X_{i,j})$,
    \item Mixed Poisson-Gaussian noise: $Y_{i,j} \sim a \mathcal{P}(X_{i,j}/a) + \mathcal{N}(0, b)$ with $(a, b) \in (\mathbb{R}^+_\ast)^2$. 
\end{itemize}

\noindent Note that in each case, $Y_{i,j}$ follows a noise model which is centered on $X_{i,j}$, \textit{i.e.} $\mathbb{E}(Y_{i,j}) = X_{i,j}$, and, as the noise is assumed to be independent at each pixel, the $Y_{i,j}$ are independent along each column. Furthermore, the $Y_{i,j}$ are also independent along each row since there are no duplicate patches in each group.

\subsection{Step 1: Unbiased risk estimate (URE)}

Recall that our objective is to get an approximation of $\Theta^\ast$ from \eqref{eq2}. While $\mathcal{R}_\Theta(X)$ is unattainable in practice, an estimate of this quantity can be calculated instead when dealing with the common types of noise given above. In any case, an unbiased estimate of the risk $\mathcal{R}_\Theta(X)$ is given by (see \Cref{appendix_nlridge_URE}):
\begin{equation}
    \begin{aligned}\operatorname{URE}_\Theta(Y) &=   \| Y\Theta - Y \|_F^2    +  2 \operatorname{tr}( D_1 \Theta ) -  \operatorname{tr}(D_1) \\
    &=2 \operatorname{tr}\left( \frac{1}{2}  \Theta^\top Q_1 \Theta  + C_1 \Theta \right)
     + \operatorname{const} \,,
    \end{aligned}
    \label{proposition1NLRidge}
\end{equation}
where $Q_1 = Y^\top Y$ is a  positive semi-definite matrix, $C_1 = D_1 - Q_1$ and $D_1$ is a diagonal matrix that depends on the type of noise:
\begin{equation}
D_1 = \left\{
     \begin{array}{ll}
        \operatorname{diag}(V^\top \mathbf{1}_n)  & \mbox{for Gaussian noise,} \\
        \operatorname{diag}(Y^\top \mathbf{1}_n)  & \mbox{for Poisson noise,} \\
        \operatorname{diag}((aY+b)^\top \mathbf{1}_n)  & \mbox{for mixed Poisson-Gaussian noise.}
    \end{array}
\right.
\end{equation}

\noindent Interestingly, \eqref{proposition1NLRidge} gives an unbiased estimate of the risk $\mathcal{R}_{\Theta}(X)$ that does not depend on $X$, but only on the observation $Y$. A common idea that has been previously exploited in image denoising, mainly for homoscedastic Gaussian noise (\textit{e.g.}, see \cite{multicomponent_sure, wavelet_sure, surelet, surenlmeans, sureGM}), is to use such an  estimate as a surrogate for minimizing the risk $\mathcal{R}_{\Theta}(X)$ in \eqref{eq2} which is inaccessible. 

\subsubsection{Minimization of the surrogate}

By minimizing \eqref{proposition1NLRidge} with respect to $\Theta$ and assuming that $Q_1 = Y^\top Y$ is positive-definite, we get the following closed-form solutions, depending on whether affine combination constraints are imposed or not (see \Cref{proposition_minURE}): 
\begin{equation}
\left\{
    \begin{array}{l}
        \displaystyle \hat{\Theta}_{lin}^{(1)}   = \mathop{\arg \min}\limits_{
\substack{\Theta \in \mathbb{R}^{k\times k}}} \operatorname{URE}_\Theta(Y)  = I_k - Q_1^{-1} D_1 \,, \\
        \displaystyle \hat{\Theta}_{aff}^{(1)} = \mathop{\arg \min}\limits_{
\substack{\Theta \in \mathbb{R}^{k\times k} \\ \text{s.t.    } \Theta^\top \mathbf{1}_k =  \mathbf{1}_k}}  \operatorname{URE}_\Theta(Y)  = I_k - \left[Q_1^{-1} - \frac{Q_1^{-1} \mathbf{1}_k (Q_1^{-1} \mathbf{1}_k)^\top}{\mathbf{1}_k^\top Q_1^{-1} \mathbf{1}_k}   \right] D_1\,.
    \end{array}
\right.
\label{theta1}
\end{equation}

\noindent In the case of conical and convex combination constraints, there exist no closed-form solution. However, by noticing that: 
\begin{equation}
\operatorname{tr}\left(\frac{1}{2} \Theta^\top Q_1 \Theta + C_1^\top \Theta \right) = \sum_{j=1}^{k} \frac{1}{2} \theta_j^\top Q_1 \theta_j + c_j^\top \theta_j \,,
\label{qp-nl_ridge}
\end{equation}
\noindent where $\theta_j$ and $c_j$ denotes the $j^{th}$ column of $\Theta$ and $C_1$, respectively, minimizing \eqref{proposition1NLRidge} under conical or convex combination constraints is nothing else that solving $k$ independent convex quadratic programming subproblems with linear constraints. Note that quadratic programs can always be solved in a finite amount of computation \cite{quadratic_programming}. Indeed, if the contents of the optimal active set (the set identifying the active constraints in the set of inequality constraints) were known in advance, we could express the active constraints as equality constraints, thereby transforming the inequality-constrained problem into a simpler equality-constrained subproblem, which in our case has a closed-form solution by exploiting the Karush–Kuhn–Tucker conditions.  Thus, if time computation were not an issue, we could iterate over all active sets (there are $2^k$ in our case, since there are $k$ inequality constraints) in the search for this optimal active set, solve the associated equality-constrained subproblem, and finally select the best solution among the $2^k$ potential candidates. Hopefully, a variety of algorithms have been developed to speed up this naive heuristic, including active-set, interior-point, or gradient projection methods \cite{quadratic_programming}. Interestingly, active-set methods solve the quadratic programming problem exactly by cleverly exploring the active sets, although they are much slower on large problems than the other methods \cite{quadratic_programming}.


In conclusion, even though conical and convex combination weights, $\hat{\Theta}_{cnl}^{(1)}$ and  $\hat{\Theta}_{cvx}^{(1)}$ respectively,  do not have closed-form expressions, they can be found exactly in a finite
amount of computation using active-set methods.

\subsubsection{On the positive definiteness of \texorpdfstring{{\boldmath$Q_1$}}{Q1}}

Positive definiteness of $Q_1$ is important to ensure the uniqueness of the minimizer of the URE \eqref{proposition1NLRidge} since $Q_1$ is the Hessian matrix of the quadratic programming subproblems defined by \eqref{qp-nl_ridge}. If $Q_1$ is positive definite, it means that the objective function is strictly convex, and as it is minimized on a convex set, the solution is unique. A priori $Q_1 = Y^\top Y$ is only positive semi-definite since for all $s \in \mathbb{R}^{k}$, $s^\top Q_1 s  = \| Y s\|_2^2 \geq 0$. However, when $n \geq k$, $Q_1$ is almost surely positive definite in general (in particular in the case of ideal additive
white Gaussian noise) as almost surely the columns of $Y$ are linearly independent. Indeed, when it is the case $s^\top Q_1 s = \| Y s\|_2^2 = 0 \Rightarrow  Y s = 0 \Rightarrow s = 0$ and $Q_1$ is then positive definite. By the way, the closed-form expressions of the combination weights \eqref{theta1} require the inverse of $Q_1$, which can be computed efficiently based on Cholesky factorization, exploiting the positive definiteness of $Q_1$ \cite{cholesky}. 

For real-world noisy images, the actual noise may deviate from the assumed ideal noise models (in general mixed Poisson-Gaussian noise). Apart from the consequences this may have on the denoising performance, an unfortunate outcome is the non positive definiteness of $Q_1$, even when $n \geq k$ (think of constant regions of the image that are, for any reason, not affected by the noise: $Y \propto \mathbf{1}_n \mathbf{1}_k^\top$). To remedy to this issue, a possible way is to consider a ``noisier'' risk compared to \eqref{riskDef} defined by:
\begin{equation}
\mathcal{R}_{\Theta}^{\text{Nr}, \alpha}(X) = \mathbb{E} \| f_\Theta(Y + \alpha W)  - X  \|_F^2\,,
\label{riskDefNr}
\end{equation}
with $\alpha > 0$ and $W \in \mathbb{R}^{n \times k}$ with $W_{i,j} \sim \mathcal{N}(0, 1)$ independent. One can prove that (see \Cref{appendix_noisier}):
\begin{equation}
\mathcal{R}_{\Theta}^{\text{Nr}, \alpha}(X) = \mathcal{R}_{\Theta}(X) + \alpha^2 n  \| \Theta \|_F^2\,,
\end{equation}
\noindent hence, an unbiased estimate of $\mathcal{R}_{\Theta}^{\text{Nr}, \alpha}(X)$ is:
\begin{equation}
    \operatorname{URE}_\Theta^{\text{Nr}, \alpha}(Y) =  \operatorname{URE}_\Theta(Y) + \alpha^2 n  \| \Theta \|_F^2 
    = 2 \operatorname{tr}\left( \frac{1}{2}  \Theta^\top (Q_1 + \alpha^2 n  I_k) \Theta  + C_1 \Theta \right) + \operatorname{const} \,,
\end{equation}
which is exactly the same expression as \eqref{proposition1NLRidge}, up to a constant, replacing $Q_1$ by $Q_1 + \alpha^2 n  I_k$. Its minimization is then given by formula \eqref{theta1} by substituting $Q_1 + \alpha^2 n  I_k$ for $Q_1$ and $D_1+n\alpha^2 I_k$ for $D_1$, respectively. The main advantage of using the ``noisier'' risk \eqref{riskDefNr} instead of the usual one \eqref{riskDef} resides in the positive definiteness of $Q_1 + \alpha^2 n  I_k$ which is always guaranteed. Indeed, for all $s \in \mathbb{R}^{k} \setminus \{0 \}$, $s^\top (Q_1 + \alpha^2 n  I_k) s  = \| Y s\|_2^2 + \alpha^2 n  \| s\|_2^2 > 0$. The choice of $\alpha$ constitutes an hyperparameter. In practice, we want $\alpha$ to be as small as possible since, for $\alpha \to 0$, the minimizer of the ``noisier'' risk \eqref{riskDefNr} converges to the minimizer of the usual risk \eqref{riskDef}.


\subsubsection{Particular case: homoscedastic Gaussian noise}

In the case of homoscedastic Gaussian noise, that is $Y_{i,j} \sim \mathcal{N}(X_{i,j}, \sigma^2)$, the expression of the URE is reduced to:
\begin{equation}
    \operatorname{SURE}_\Theta(Y) = \| Y\Theta - Y \|_F^2    +  2 n \sigma^2  \operatorname{tr}(\Theta) - nk\sigma^2\,,
\end{equation}
which is nothing else than Stein's unbiased risk estimate \cite{SURE}. Considering unconstrained minimization, the estimated optimal weights are:
\begin{equation}
\hat{\Theta}_{lin}^{(1)} = \arg \min_{\Theta \in \mathbb{R}^{k \times k}} \operatorname{SURE}_\Theta(Y)  = I_k - n\sigma^2 (Y^\top Y)^{-1}\,.
\end{equation}

\noindent Note that $\hat{\Theta}^{(1)}$ is close to $\Theta^{\ast}$ as long as the variance of SURE is low. A rule of thumbs used in \cite{surelet} states that the number of parameters must not be ``too large'' compared to the number of data in order for the variance of SURE to remain small. In our case, this suggests that $n > k$. This result suggests that NL-Ridge is expected to be efficient during this step if a few large patches are used. This is consistent with the condition for which $Q_1 = Y^\top Y$ is almost surely positive definite.

\subsection{Step 2: Internal adaptation}
\label{internal_adaptation_nl_ridge_section}

At the end of Step 1, we get a first denoised image $\hat{\mathcal{I}}^{(1)}$ which will serve as a pilot in the second step. Once again, we focus on the solution of \eqref{eq2} to denoise locally similar patches. As $X$ and $\hat{X}^{(1)}$, the corresponding group of similar patches in $\hat{\mathcal{I}}^{(1)}$, are supposed to be close, the ``internal adaptation'' procedure \cite{LIDIA} consists in solving \eqref{eq2} by substituting $\hat{X}^{(1)}$ for $X$. 

Interestingly, for any of the types of noise studied, the quadratic risk \eqref{riskDef} has a closed-form expression (see \Cref{lemma3}):
\begin{equation}
    \mathcal{R}_\Theta(X) = \| X\Theta - X \|^2_F  + \operatorname{tr}(\Theta^\top D_2 \Theta)
  =2 \operatorname{tr}\left( \frac{1}{2}  \Theta^\top Q_2 \Theta  + C_2 \Theta \right)
     + \operatorname{const} \,,
     \label{proposition3NLRidge}
\end{equation}
where $Q_2 = X^\top X + D_2$ is a  positive semi-definite matrix, $C_2 = D_2 - Q_2$ and $D_2$ is a diagonal matrix that depends on the type of noise:
\begin{equation}
D_2 = \left\{
     \begin{array}{ll}
        \operatorname{diag}(V^\top \mathbf{1}_n)  & \mbox{for Gaussian noise,} \\
        \operatorname{diag}(X^\top \mathbf{1}_n)  & \mbox{for Poisson noise,} \\
        \operatorname{diag}((aX+b)^\top \mathbf{1}_n)  & \mbox{for mixed Poisson-Gaussian noise.}
    \end{array}
\right.
\end{equation}

\noindent Substituting $\hat{X}^{(1)}$ for $X$ in expression \eqref{proposition3NLRidge}, a natural surrogate for  $\mathcal{R}_\Theta(X)$ is  $\mathcal{R}_\Theta(\hat{X}^{(1)})$.
A second approximation of \eqref{eq2} can then be deduced by minimizing this latter. Interestingly, this second estimate $\hat{\Theta}^{(2)}$
produces a significant boost in terms of denoising performance compared to $\hat{\Theta}^{(1)}$. Even if the second step can be iterated but we did not notice improvements in our experiments.

\subsubsection{Minimization of the surrogate}

By minimizing \eqref{proposition3NLRidge} where $X$ is replaced by $\hat{X}^{(1)}$ with respect to $\Theta$ and assuming that $Q_2$ is positive-definite, we get the following closed-form solutions, depending on whether affine combination constraints are imposed or not (see \Cref{prop_nlridge_minimisation_risk}): 
\begin{equation}
\left\{
    \begin{array}{l}
      \displaystyle \hat{\Theta}_{lin}^{(2)} = \mathop{\arg \min}\limits_{
\substack{\Theta \in \mathbb{R}^{k\times k}}} \mathcal{R}_\Theta(\hat{X}^{(1)})  = I_k - Q_2^{-1} D_2  
\,,  \\
\displaystyle \hat{\Theta}_{aff}^{(2)} = \mathop{\arg \min}\limits_{
\substack{\Theta \in \mathbb{R}^{k\times k} \\ \text{s.t.    } \Theta^\top \mathbf{1}_k =  \mathbf{1}_k}}  \mathcal{R}_\Theta(\hat{X}^{(1)})  = I_k - \left[Q_2^{-1} - \frac{Q_2^{-1} \mathbf{1}_k (Q_2^{-1} \mathbf{1}_k)^\top}{\mathbf{1}_k^\top Q_2^{-1} \mathbf{1}_k}   \right] D_2\,.
    \end{array}
\right.
\label{theta2}
\end{equation}

\noindent As in Step 1, there is no closed-form solution in the case of conical and convex combination constraints. $\hat{\Theta}_{cnl}^{(2)}$ and $\hat{\Theta}_{cvx}^{(2)}$ can however be approximated using any of iterative algorithms \cite{quadratic_programming} dedicated to the resolution of convex quadratic programming problems.

\subsubsection{On the positive definiteness of \texorpdfstring{{\boldmath$Q_2$}}{Q2}}

Compared to Step 1, $Q_2 = \hat{X}^{(1) \top} \hat{X}^{(1)} + D_2$ is much more likely to be positive definite. In fact, as soon as $D_2$ has positive diagonal elements, which is always the case except for Poisson noise with $\hat{X}^{(1)} = 0$, $Q_2$ is positive definite. But this case can be treated separately by setting arbitrarily $\hat{\Theta}^{(2)} = I_k$ for example.

\subsubsection{Particular case: homoscedastic Gaussian noise}

In the case of homoscedastic Gaussian noise, that is $Y_{i,j} \sim \mathcal{N}(X_{i,j}, \sigma^2)$, the expression of the quadratic risk is reduced to:
\begin{equation}
    \mathcal{R}_\Theta(X) = \| X\Theta - X \|^2_F  + n\sigma^2 \| \Theta \|^2_F\,,
    \label{risk_homo2}
\end{equation}
which is nothing else than the expression of a multivariate Ridge regression. Considering unconstrained minimization, the estimated optimal weights are then:
\begin{equation}
\hat{\Theta}_{lin}^{(2)} = \arg \min_{\Theta \in \mathbb{R}^{k \times k}} \mathcal{R}_\Theta(\hat{X}^{(1)}) = I_k - n\sigma^2 \left(\hat{X}^{(1) \top} \hat{X}^{(1)} + n\sigma^2 I_k \right)^{-1}  \,.
\end{equation}

\noindent It is interesting to compare the behavior of the weights $\hat{\Theta}_{lin}^{(2)}$ and $\hat{\Theta}_{aff}^{(2)}$ when $\sigma$ tends to  $+\infty$. In fact, the higher $\sigma$ and the more important the second term $n\sigma^2 \| \Theta \|^2_F$ is in the expression of the risk \eqref{risk_homo2} (and the less the dependence on $X$). At the limit, when $\sigma \to +\infty$:
\begin{equation}
\left\{
    \begin{array}{l}
       \displaystyle \hat{\Theta}_{lin}^{(2)} = \mathop{\arg \min}\limits_{
\substack{\Theta \in \mathbb{R}^{k\times k}}}  \mathcal{R}_\Theta(X)  \to \mathbf{0}_k \mathbf{0}_k^\top
\,, \\
        \displaystyle \hat{\Theta}_{aff}^{(2)} = \mathop{\arg \min}\limits_{
\substack{\Theta \in \mathbb{R}^{k\times k} \\ \text{s.t.    } \Theta^\top \mathbf{1}_k =  \mathbf{1}_k}}  \mathcal{R}_\Theta(\hat{X}^{(1)}) \to \mathbf{1}_k \mathbf{1}_k^\top / k\,.
    \end{array}
\right.
\end{equation}

\noindent As a consequence, the final produced image $\hat{\mathcal{I}}^{(2)}$ tends towards the ``zero-image'' in the case of unconstrained minimization, whereas, in the case of affine combination of patches, $\hat{\mathcal{I}}^{(2)}$ consists of simple averages of similar patches.  This fundamental difference in the asymptotic behavior of the weights may explain why affine patch combinations are more recommended when the noise level increases.

\subsection{Weighted average reprojection}

After the denoising of a group of similar patches, each denoised patch is repositioned at its right location in the image. As several pixels are denoised multiple times, a final step of  aggregation, or reprojection \cite{Aggreg}, is necessary to produce a final denoised image $\hat{\mathcal{I}}^{(1)}$ or $\hat{\mathcal{I}}^{(2)}$. With inspiration from \cite{Aggreg}, each pixel belonging to column $j$ of $Y$ is assigned, after denoising, the weight $w_j  = 1 / ( \|\Theta_{\cdot,j} \|_2^{2})$. Those weights are at the end pixel-wise normalized such that the sum of all weights associated to a same pixel equals one. 

\bigskip

The complete NL-Ridge method for image denoising is summarized in  \Cref{algo_nlridge}. Please note the difference between a freshly denoised similarity matrix, denoted $\check{X}_g$, and its aggregated equivalent $\hat{X}_g$.

\begin{algorithm}
\caption{NL-Ridge algorithm for image denoising}
\label{algo_nlridge}
\begin{algorithmic}
\STATE{\textbf{Input:} Noisy image $y$, patch and group sizes for step 1 and step 2: $\sqrt{n_1}$, $\sqrt{n_2}$, $k_1$ and $k_2$}

\STATE{\texttt{/* Step 1}}
\FOR{each $\sqrt{n_1} \times \sqrt{n_1}$ overlapping noisy patch in $y$}
\STATE{Extract its $k_1$ most similar patches to form similarity matrix $Y_g^{(1)}$}
\STATE{Estimate combination weights $\hat{\Theta}_g^{(1)}$ with formula \eqref{theta1} \textit{(closed-form expression)}}
\STATE{Perform collaborative denoising $\check{X}^{(1)}_g = Y_g ^{(1)} \hat{\Theta}_g^{(1)}$}
\ENDFOR

\STATE{Aggregate all the denoised patches contained in the groups $\check{X}^{(1)}_g$ to form the estimated image $\hat{\mathcal{I}}^{(1)}$}

\STATE{\texttt{/* Step 2}}
\FOR{each $\sqrt{n_2} \times \sqrt{n_2}$ overlapping patch in $\hat{\mathcal{I}}^{(1)}$}
\STATE{Extract its $k_2$ most similar patches in $\hat{\mathcal{I}}^{(1)}$ to form similarity matrix $\hat{X}^{(1)}_g$}
\STATE{Extract the  $k_2$ corresponding noisy patches in $y$ to form similarity matrix $Y_g^{(2)}$}
\STATE{Estimate combination weights $\hat{\Theta}_g^{(2)}$ with  \eqref{theta2}  \textit{(closed-form expression)}}
\STATE{Perform collaborative denoising  $\check{X}^{(2)}_g = Y_g^{(2)} \hat{\Theta}_g^{(2)}$}
\ENDFOR

\STATE{Aggregate all the denoised patches contained in the groups $\check{X}^{(2)}_g$ to form the estimated image $\hat{\mathcal{I}}^{(2)}$}

\RETURN{$\hat{\mathcal{I}}^{(2)}$}
\end{algorithmic}
\end{algorithm}
\section{A unified view of non-local denoisers}
\label{nlridge_section3}

In NL-Ridge, the local denoiser $f_\Theta$ is arbitrarily of the form given by \eqref{local_denoiser0} involving the linear combinations of similar patches with closed-form aggregation weights given in \eqref{theta1} and \eqref{theta2} for unconstrained minimization. In this section, we show that NL-Ridge can serve to interpret two popular state-of-the-art non-local methods - NL-Bayes \cite{nlbayes} and BM3D \cite{BM3D} - which were originally designed with two very different modeling and estimation frameworks. It amounts actually to considering two particular families $(f_\Theta)$ of local denoisers. In the rest, we focus exclusively on homoscedastic Gaussian noise, that is $Y_{i,j} \sim \mathcal{N}(X_{i,j}, \sigma^2)$. All the proofs of this section can be found in \Cref{sectionNLBayes} and \Cref{sectionBM3D}.

\subsection{Analysis of NL-Bayes algorithm}

The NL-Bayes \cite{nlbayes} algorithm has been established in the Bayesian setting and the resulting maximum a posteriori estimator is computed with a two-step procedure as NL-Ridge. Adopting a novel parametric view of this algorithm, let us consider the following family of local denoisers as starting point:
\begin{equation}
f_{\Theta, \beta} : Y \in \mathbb{R}^{n \times k} \mapsto  \Theta Y + \beta \mathbf{1}_k^\top
\label{family_nlbayes}
\end{equation}
\noindent where $\Theta \in \mathbb{R}^{n \times n}$ and $\beta \in \mathbb{R}^{n}$. Our objective is to find $(\Theta^\ast, \beta^\ast)$ that minimizes the quadratic risk $\mathcal{R}_{\Theta, \beta}(X)  = \mathbb{E} \| f_{\Theta, \beta}(Y)- X \|^2_F$, that is:
\begin{equation}
    \Theta^\ast, \beta^\ast = \arg \min_{\Theta, \beta} \mathcal{R}_{\Theta, \beta}(X)\,.
\end{equation}

\subsubsection{Step 1: Unbiased risk estimate (URE)}

 In the case of Gaussian noise, Stein's unbiased  estimate of the quadratic risk $\mathcal{R}_{\Theta, \beta}(X) = \mathbb{E} \| f_{\Theta, \beta}(Y)  - X  \|_F^2$ is:
\begin{equation}
    \operatorname{SURE}_{\Theta, \beta}(Y) = \| \Theta Y  - Y  + \beta \mathbf{1}_k^\top \|_F^2  + 2 k \sigma^2 \operatorname{tr}(\Theta) - nk\sigma^2\,,
\end{equation}
which reaches its minimum for:
\begin{equation}
    \hat{\Theta}^{(1)} = (C_Y - \sigma^2 I_n)C_Y^{-1} \quad \text{and} \quad  \hat{\beta}^{(1)}  = (I_n - \hat{\Theta}^{(1)} ) \mu_Y\,,
\end{equation}
\noindent where $\mu_Y \in \mathbb{R}^{k}$ and $C_Y \in \mathbb{R}^{n \times n}$ denote the empirical mean and covariance matrix of a group of patches $Y \in \mathbb{R}^{n \times k}$, that is 
\begin{equation}\mu_Y = \frac{1}{k} Y \mathbf{1}_k 
\quad  \text{and} \quad 
C_Y  = \frac{1}{k} (Y  - \mu_Y \mathbf{1}_k^\top )  (Y  - \mu_Y \mathbf{1}_k^\top )^\top   \,. \end{equation}
\noindent  Interestingly, $f_{\hat{\Theta}^{(1)}, \hat{\beta}^{(1)}}(Y)$ is the expression given in \cite{nlbayes} (Step 1), which is actually derived from the prior distribution of patches assumed to be Gaussian. Furthermore, our framework provides guidance on the choice of the parameters $n$ and $k$. Indeed, SURE is helpful provided that its variance remains small which is achieved if $n < k$ (the number of parameters must not be “too large” compared to the number of data). This result suggests that NL-Bayes is expected to be efficient if small patches are used, as confirmed in the experiments in \cite{nlbayesImp}.

\subsubsection{Step 2: ``Internal adaptation''}

The quadratic risk $\mathcal{R}_{\Theta, \beta}(X)$ associated with the family of functions defined in \eqref{family_nlbayes} has a closed-form expression:
\begin{equation}
     \mathcal{R}_{\Theta, \beta}(X) = \|   \Theta X   - X +  \beta \mathbf{1}_k^\top \|_F^2 + k\sigma^2 \| \Theta \|_F^2\,.
\end{equation}

\noindent Interpreting the second step in \cite{nlbayes} as an ``internal adaptation'' step, we want to minimize the risk $\mathcal{R}_{\Theta, \beta}(X)$ by substituting $\hat{X}^{(1)}$, obtained at the end of step 1, for $X$, which is unknown. The updated parameters become: 
\begin{equation}
    \hat{\Theta}^{(2)} = C_{\hat{X}^{(1)}}(C_{\hat{X}^{(1)}} + \sigma^2 I_n)^{-1} \quad \text{and} \quad  \hat{\beta}^{(2)}  = (I_n - \hat{\Theta}^{(2)})\mu_{\hat{X}^{(1)}}\,,
\end{equation}
and $f_{\hat{\Theta}^{(2)}, \hat{\beta}^{(2)}}(Y)$ corresponds to the original second-step expression in \cite{nlbayes}. 

\subsection{Analysis of BM3D algorithm}

 BM3D \cite{BM3D} is probably the most popular non-local method for image denoising. It assumes a locally sparse representation of images in a transform domain. A two step algorithm  was described in \cite{BM3D} to achieve state-of-the-art results for several years. By using the generic  NL-Ridge  formulation, we consider the following family of functions:
\begin{equation}
f_{\Theta} : Y \mapsto  P^{-1} (\Theta \odot (P Y Q)) Q^{-1}
\label{bm3d_family}
\end{equation}
\noindent where $\Theta \in \mathbb{R}^{n \times m}$ and where $P \in \mathbb{R}^{n \times n}$ and $Q \in \mathbb{R}^{m \times m}$ are two orthogonal matrices that model a separable 3D-transform (typically a 2D and 1D \textit{Discrete Cosine Transform}, respectively). Once again, our objective is to find $\Theta^\ast$ that minimizes the quadratic risk $\mathcal{R}_{\Theta}(X) =\mathbb{E} \| f_\Theta(Y) - X \|_F^2$, that is:
\begin{equation}
    \Theta^\ast = \arg \min_{\Theta} \mathcal{R}_{\Theta}(X)\,.
\end{equation}

\subsubsection{Step 1: Unbiased risk estimate (URE)}

Stein's unbiased risk estimate (SURE) is defined as follows:  
\begin{equation}
\operatorname{SURE}_{\Theta}(Y) = \|  (\Theta - \mathbf{1}_n \mathbf{1}_k^\top) \odot  PYQ \|_F^2  +2\sigma^2 \langle \Theta, \mathbf{1}_n \mathbf{1}_k^\top \rangle_F -nk\sigma^2\,,
\end{equation}
and its minimization yields:
\begin{equation}
    \hat{\Theta}^{(1)}_a = \mathbf{1}_n  \mathbf{1}_k^\top - \frac{\sigma^2}{(PYQ)^{\odot2}} \,,
\end{equation}

\noindent where the division is element-wise. Unfortunately, $f_{\hat{\Theta}^{(1)}_{a}}(Y)$ does not provide very satisfying denoising results. This result is actually expected as the number of parameters equals the size of data ($n \times k$), making SURE weakly efficient. To overcome this difficulty, we can force the elements of $\Theta$ to be either $0$ or $1$, \textit{i.e.} by imposing the search space to be $\Theta \in \{0, 1\}^{n \times k}$. Minimizing SURE under this constraint results in the alternative estimation:
\begin{equation}
    \hat{\Theta}^{(1)}_{b} =\mathds{1}_{\mathbb{R} \setminus [-\sqrt{2} \sigma, \sqrt{2} \sigma]}(PYQ)\,.
\end{equation}
 $f_{\hat{\Theta}^{(1)}_{b}}(Y)$ acts then as a hard thresholding estimator as in BM3D: the coefficients of the transform domain (\textit{i.e.} the elements of the matrix $PYQ$) below  $\sqrt{2}\sigma$, in absolute value, are canceled before applying the inverse 3D-transform. This result suggests that the threshold should be linearly dependent on $\sigma$ but also that the threshold value is independent on the choice of the orthogonal transforms $P$ and $Q$. In \cite{BM3D}, a threshold value   of $2.7\sigma$ was carefully chosen in Step 1, which is approximately twice the SURE-prescribed threshold.

\subsubsection{Step 2: ``Internal adaptation''}

The quadratic risk $\mathcal{R}_\Theta(X)$ associated with the family of functions defined in \eqref{bm3d_family} has a closed-form
expression:
\begin{equation}
    \mathcal{R}_{\Theta}(X) = \mathbb{E} \| f_{\Theta}(Y) - X \|_F^2 = \|  (\Theta - \mathbf{1}_n \mathbf{1}_k^\top) \odot PXQ \|_F^2 + \sigma^2 \| \Theta \|_F^2 \,,
    \label{riskBM3D_}
\end{equation}

\noindent and the  ``internal adaptation'' step yields the same expression as the Wiener filtering used in Step 2 in BM3D \cite{BM3D}: 
\begin{equation}
    \hat{\Theta}^{(2)} = \frac{(P\hat{X}^{(1)}Q)^2}{\sigma^2 + (P\hat{X}^{(1)} Q)^{\odot2}}\,.
\end{equation}

\noindent  However, it is worth noting that the closed-form expression of the risk \eqref{riskBM3D_} is obtained by assuming that the coefficients of $Y$ are all independent. Thus, theoretically, $Y$ should gather together exclusively non-overlapping patches. This important limitation is not yet considered in the original paper \cite{BM3D}. Hopefully, this has little effect on the denoising performance. More recently, a new version of the algorithm has been published that takes into account when the noise in one patch is correlated with the noise in one of the other patches \cite{BM3Derratum}.

\bigskip 

\begin{figure}[t]
    \centering
    \begin{tabular}[t]{ccc}
             
             \includegraphics[width=0.2\columnwidth, valign=m, trim = 0px 50px  720px 40px , clip]{./Figures/BM2.pdf}
            
          &  \includegraphics[width=0.5\columnwidth, valign=m, trim = 120px 0px  130px 0px , clip]{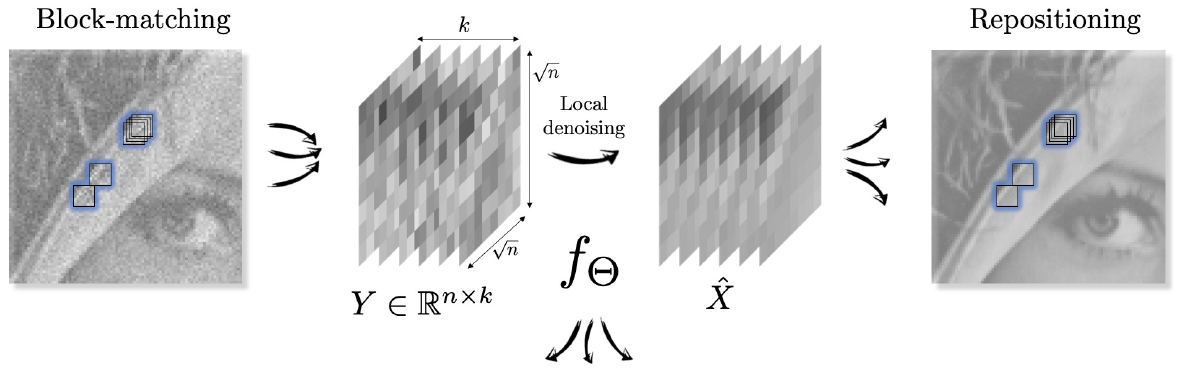} &

          \includegraphics[width=0.2\columnwidth, valign=m, trim = 720px 50px 0px 40px , clip]{./Figures/BM2.pdf}
    \end{tabular}

    \renewcommand{\arraystretch}{1.2}
    
\footnotesize 
\begin{tabularx}{\textwidth}{XXX}
    \textbf{BM3D} \cite{BM3D} assumes a locally sparse representation in a transform domain: &  \textbf{NL-Bayes} \cite{nlbayes} was originally established in the Bayesian setting: &   \textbf{NL-Ridge} (ours) leverages linear combinations of noisy patches:
\end{tabularx}

\begin{tabularx}{\textwidth}{YYY}
  $f_{\Theta}(Y) =  P^\top (\Theta \odot (P Y Q)) Q^\top\,,$ & $f_{\Theta, \beta}(Y) = \Theta Y + \beta \mathbf{1}_k^\top\,,$ & $\boxed{f_{\Theta}(Y) =  Y \Theta}$\,.
\end{tabularx}

\begin{tabularx}{\textwidth}{XXX}
    $P, Q$: orthogonal matrices. & $\mathbf{1}_k$: $k$-dimensional all-ones vector. &
\end{tabularx}

    \caption[Proposed unifying parametric framework for unsupervised non-local denoisers]{Illustration of the parametric view of several popular non-local denoisers. Examples of parameterized functions unequivocally identifying the denoiser are given, whose optimal parameters are eventually selected for each group of patches by ``internal adaptation'' optimization.} 
    \label{bm00}
\end{figure}

In summary, we have shown that BM3D \cite{BM3D} and NL-Bayes \cite{nlbayes} can be interpreted under a parametric view within NL-Ridge framework in the case of homoscedastic Gaussian noise. \Cref{bm00} summarizes the three different algorithms which are distinguished by their parametric families. Our novel paradigm has some advantages beyond the unification of methods: it enables to set the size of the patches and may potentially relax the need to specify the prior distribution of patches.

\section{Experimental results}
\label{nlridge_section4}

\addtolength{\tabcolsep}{-5pt} 
\begin{figure*}[t]
\centering
\renewcommand{\arraystretch}{0.5}
\begin{tabular}{cc}

\begin{tabular}{c}
\begin{tikzpicture}
\node[anchor=south west,inner sep=0] (image) at (0,0)
  {\includegraphics[width=0.252\textwidth]{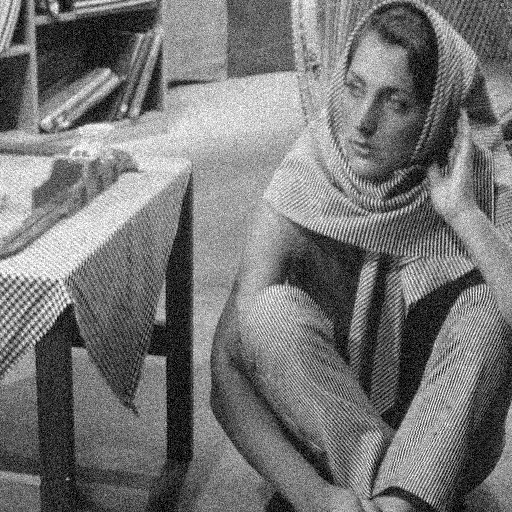}};
  

\node at (image.south east) [anchor=south east, inner sep=0] {\includegraphics[width=0.147\textwidth, trim=100px 280px 308px 180px, clip]{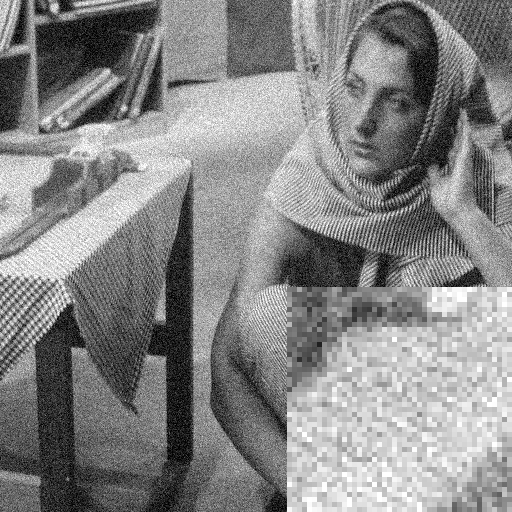}};

\begin{scope}[x={(image.south east)},y={(image.north west)}]
  \begin{scope}[shift={(0,1)},x={(1/512,0)},y={(0,-1/512)}]
  
   \draw[dodgerblue, semithick] (100, 180) rectangle (204, 232);
   
   \draw[dodgerblue, thick] (212, 362) rectangle (512, 512);

    \draw[densely dashed, thin, dodgerblue] (100, 232) -> (212, 512);
    
    \draw[densely dashed, thin, dodgerblue] (204, 180) -> (512, 362);
  \end{scope}
\end{scope}
\end{tikzpicture} \\
\scriptsize Noisy 
\end{tabular}
& 
\begin{tabular}{ccc}
\includegraphics[width=0.24\textwidth, trim=100px 280px 308px 180px, clip]{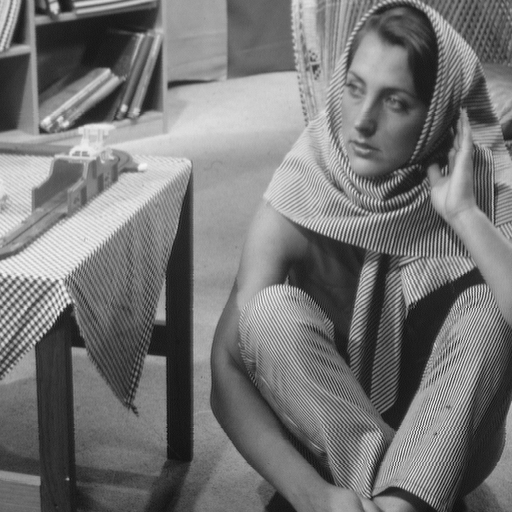} & \includegraphics[width=0.24\textwidth, trim=100px 280px 308px 180px, clip]{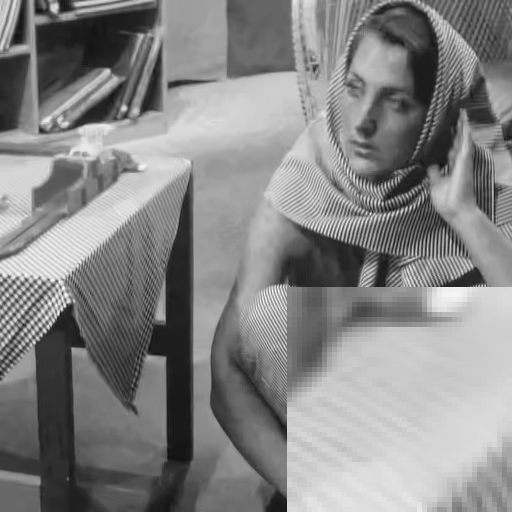} & \includegraphics[width=0.24\textwidth, trim=100px 280px 308px 180px, clip]{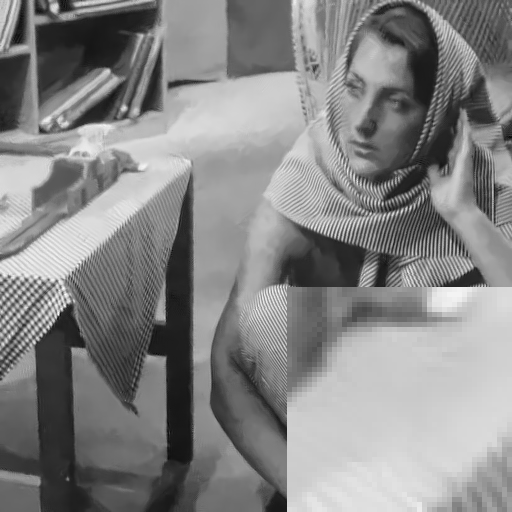}  \\
\scriptsize Ground truth   &   \scriptsize BM3D \cite{BM3D} / 31.72 dB & \scriptsize NL-Bayes \cite{nlbayes} / 31.54 dB   \\

\includegraphics[width=0.24\textwidth, trim=100px 280px 308px 180px, clip]{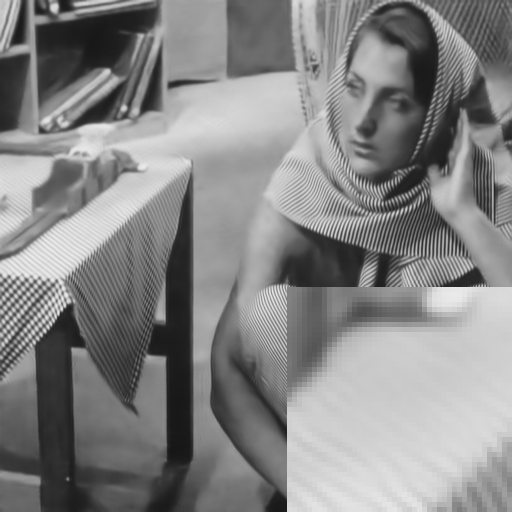} &
\includegraphics[width=0.24\textwidth, trim=100px 280px 308px 180px, clip]{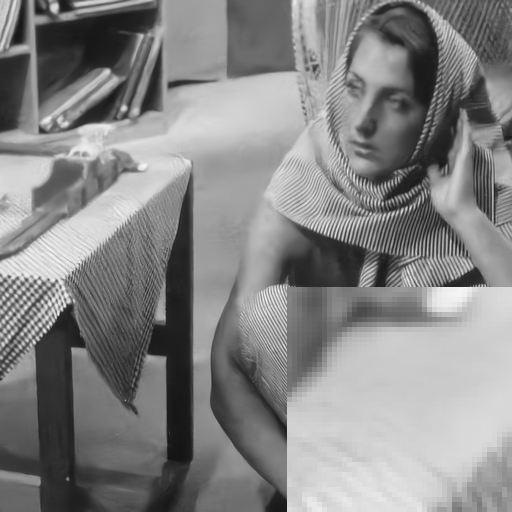}  & 
\includegraphics[width=0.24\textwidth, trim=100px 280px 308px 180px, clip]{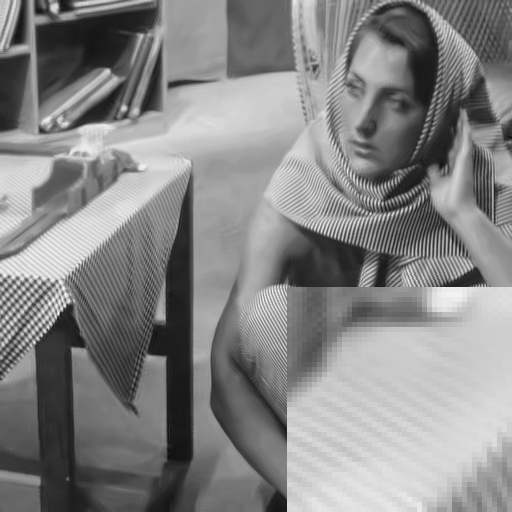} \\

\scriptsize Self2Self \cite{S2S} / 31.62 dB   &
\scriptsize DnCNN \cite{dncnn} / 31.06 dB  & \scriptsize NL-Ridge (ours) / \textbf{32.06} dB \\
\end{tabular}
\end{tabular}

\caption[Qualitative comparison of image denoising results with synthetic white Gaussian noise]{Denoising results (in PSNR) on \textit{Barbara} corrupted with additive white Gaussian noise ($\sigma = 20$).}
\label{nlridge_photo}
\end{figure*}
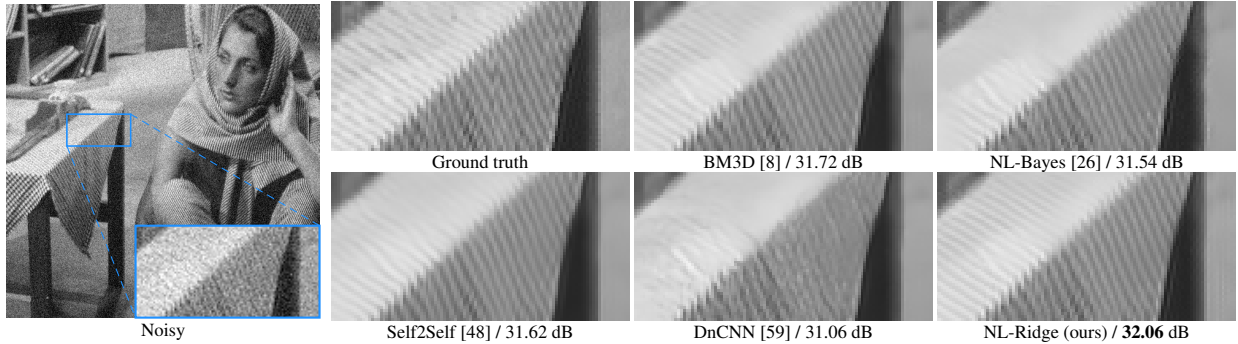
\addtolength{\tabcolsep}{5pt}

In this section, we compare the performance of our NL-Ridge method with state-of-the-art methods, including related network-based methods \cite{dncnn, ffdnet, LIDIA, S2S, N2S, DIP, scunet, drunet, restormer, N2F, ZS-N2N, rethinking} applied to standard gray images artificially corrupted with homoscedastic Gaussian noise with zero mean and variance $\sigma^2$ and on real-world noisy images, modeled by mixed Poisson-Gaussian noise. We used the implementations provided by the authors as well as the corresponding trained weights for supervised networks. Performances of NL-Ridge and other methods are assessed in terms of PSNR values when the ground truth is available. Unless specified, NL-Ridge is run without constraint on the weights of the linear combinations. The code can be downloaded at: \href{https://github.com/sherbret/NL-Ridge/}{https://github.com/sherbret/NL-Ridge/}.

\subsection{Setting of algorithm parameters}

For the sake of computational efficiency, the search for similar patches, computed in the $\ell_2$ sense, across the image is restricted to a small local window $\kappa \times \kappa$ centered around each reference patch (in our experiments $\kappa=37$). Considering iteratively each overlapping patch of the image as reference patch is also computationally demanding, therefore only an overlapping patch over $\delta$, both horizontally and vertically, is considered as a reference patch. The number of reference patches and thus the time spent searching for similar patches is then divided by $\delta^2$. This common technique \cite{BM3D, nlridge, WNNM} is sometimes referred in the literature as the \textit{step trick}. In our experiments, we take $\delta = 4$.

\begin{table}[t]
\footnotesize
\caption{Setting of algorithm parameters (patch size $n$, patch number $k$) depending on the noise standard deviation $\sigma$ values.}
\centering
\begin{tabular}{ccccc}
  \hline
   $\sigma$ & $n_{1}$ & $n_{2}$ & $k_{1}$ & $k_{2}$ \\\hline \hline  
 $\textcolor{white}{1}0< \sigma \leq 15$ & $7\times7$ & $7\times7$ & 18 & 55\\\hline
 $15 < \sigma \leq 35$ & $9\times9$ & $9\times9$ & 18 & 90\\\hline
 $35 < \sigma \leq 50$ & $11\times11$ & $9\times9$ & 20 & 120\\\hline
\end{tabular}
\label{nlridge_optimalParams}
\end{table}

Finally, the choice of the parameters $n$ and $k$ depend on the noise level. Experimentally, bigger patches have to be considered for higher noise levels as well as a higher quantity of patches for the second step. An empirical analysis led us to choose the parameters reported in  \Cref{nlridge_optimalParams} for homoscedastic Gaussian noise of variance $\sigma^2$.

\afterpage{%
    \clearpage
    \thispagestyle{empty}

\begin{landscape}

\begin{table*}[!t]
\centering
\caption{The PSNR (dB) results of different methods applied to three datasets corrupted with synthetic white Gaussian noise. The best method among each category (dataset-based or single-image) is emphasized in bold.}


\resizebox{\columnwidth}{!}{%
  \begin{NiceTabular}{c@{\hspace{0.1cm}} c@{\hspace{0.1cm}} c c@{\hspace{0.5cm}}
c@{\hspace{0.08cm}} c@{\hspace{0.08cm}} c@{\hspace{0.08cm}} c@{\hspace{0.08cm}} c@{\hspace{0.08cm}} c@{\hspace{0.08cm}} c@{\hspace{0.08cm}} c@{\hspace{0.08cm}} c@{\hspace{0.7cm}}
c@{\hspace{0.08cm}} c@{\hspace{0.08cm}} c@{\hspace{0.08cm}} c@{\hspace{0.08cm}} c@{\hspace{0.08cm}} c@{\hspace{0.08cm}} c@{\hspace{0.08cm}} c@{\hspace{0.08cm}} c@{\hspace{0.7cm}}
c@{\hspace{0.08cm}} c@{\hspace{0.08cm}} c@{\hspace{0.08cm}} c@{\hspace{0.08cm}} c@{\hspace{0.08cm}} c@{\hspace{0.08cm}} c@{\hspace{0.08cm}} c@{\hspace{0.08cm}} c@{\hspace{0.0cm}} c}
  \hline
 &&&  \textbf{Methods}   & \multicolumn{9}{c}{\textbf{Set12}} & \multicolumn{9}{c}{\textbf{BSD68}} & \multicolumn{9}{c}{\textbf{Urban100}} & \\\hline\hline\noalign{\vskip 0.1cm}

      &&& Noise level $\sigma$ & 5 &/&  15 &/& 25 &/&  35 &/& 50 & 5 &/&  15 &/& 25 &/&  35 &/& 50
      & 5 &/&  15 &/& 25 &/&  35 &/& 50
      & \\[0.1cm]

   \hline\noalign{\vskip 0.1cm}

\multirow{6}{*}{\begin{sideways} \scriptsize \textbf{Dataset-based}  \end{sideways}} & \multirow{6}{*}{\begin{sideways} \scriptsize \textit{Neural}  \end{sideways}} & \multirow{6}{*}{\begin{sideways} \scriptsize \textit{network}  \end{sideways}} & DnCNN \cite{dncnn} &  37.74 &/& 32.86 &/& 30.44 &/& 28.82 &/& 27.18 & 37.71 &/& 31.73 &/& 29.23 &/& 27.69  &/& 26.23 & 37.52 &/& 32.68 &/& 29.97 &/& 28.11 &/& 26.28  & \\
 &&& FFDNet \cite{ffdnet} &  38.11 &/& 32.75 &/& 30.43 &/& 28.92 &/& 27.32 & 37.80 &/& 31.63 &/& 29.19 &/& 27.73 &/& 26.29 &  38.12 &/& 32.43 &/& 29.92 &/& 28.27 &/& 26.52 & \\
 &&& LIDIA \cite{LIDIA} &  - &/& 32.85 &/& 30.41 &/& -&/&  27.19 & - &/& 31.62 &/& 29.11 &/& -&/&  26.17 & - &/& 32.80 &/& 30.12 &/& - &/& 26.51 & \\


 &&& DRUNet \cite{drunet} & 38.64  &/& 33.25 &/&  30.94 &/& 29.45 &/&  27.90 & 38.07 &/& 31.91  &/& 29.48 &/& 28.00 &/&  26.59 & 38.91 &/& 33.44 &/& 31.11  &/& 29.61 &/& 27.96  & \\
 &&& Restormer \cite{restormer} & \bf 38.70  &/& 33.42 &/& 31.08  &/& \bf 29.57 &/&  28.01 & \bf 38.11 &/& 31.96  &/& 29.52  &/& \bf 28.05 &/&  26.62 & \bf 39.06 &/& 33.79 &/& 31.46  &/& \bf 30.00 &/& 28.33 & \\
 &&& SCUNet \cite{scunet} & -  &/& \bf 33.43 &/& \bf 31.09 &/& - &/&  \bf 28.04 & - &/& \bf 31.99 &/& \bf 29.55 &/& - &/&  \bf 26.67 & - &/& \bf 33.88 &/& \bf 31.58 &/& - &/& \bf 28.56 & \\

 \hline\noalign{\vskip 0.1cm}

 \multirow{15}{*}{\begin{sideways} \scriptsize \textbf{Single-image} \end{sideways}} & \multirow{6}{*}{\begin{sideways} \scriptsize \textit{Neural} \end{sideways}} & \multirow{6}{*}{\begin{sideways} \scriptsize \textit{network} \end{sideways}} & DIP \cite{DIP}  &  -  &/&  30.12  &/& 27.54 &/& - &/& 24.67  &  - &/& 28.83 &/& 26.59 &/& - &/& 24.13  & - &/& - &/& -  &/& - &/& - &\\
 &&& N2S \cite{N2S}    &  - &/& 31.01 &/& 28.64 &/& - &/& 25.30 & - &/&  29.46 &/& 27.72 &/& - &/& 24.77  & - &/& - &/& -  &/& - &/& - &\\

  &&& N2F \cite{N2F}   &  - &/& 30.21 &/& 28.17 &/& - &/& 25.09 &  - &/& 29.82 &/& 27.79 &/& -&/&  25.05 & - &/& - &/& -  &/& - &/& - &\\

 &&& ZS-N2N \cite{ZS-N2N}   &  - &/& 30.05 &/& 27.26 &/& - &/& 23.56 &  - &/& 29.61 &/& 26.86 &/&  - &/&  23.46 & - &/& - &/& -  &/& - &/& - &\\

 &&& S2S \cite{S2S}   &  - &/& 32.07 &/& \bf 30.02 &/& - &/& 26.49 &  - &/& 30.62 &/& 28.60 &/& -&/&  25.70 & - &/& - &/& -  &/& - &/& - &\\ 

  &&& RDIP \cite{rethinking}   &  - &/& \bf 32.26 &/& 29.79 &/& - &/& \bf 26.60 &  - &/& \bf 31.21 &/& \bf 28.78 &/& -&/&  \bf 25.81 & - &/& - &/& -  &/& - &/& - &\\

 \cdashline{2-32}\noalign{\vskip 0.1cm}

 & \multirow{6}{*}{\begin{sideways} \scriptsize \textit{Traditional} \end{sideways}} & \multirow{6}{*}{\begin{sideways} \scriptsize \textit{2-step} \end{sideways}} & BM3D \cite{BM3D}  &  38.02 &/& 32.37 &/& 29.97 &/& 28.40 &/& 26.72 &  37.55 &/& 31.07 &/& 28.57 &/& 27.08 &/& 25.62 & 38.30 &/& 32.35  &/& 29.70 &/& 27.97 &/&  25.95 &\\
 &&& NL-Bayes \cite{nlbayes} &  38.12 &/&  32.25 &/& 29.88 &/& 28.30 &/&  26.45 & 37.62 &/& 31.16 &/& \bf 28.70 &/&  \bf 27.18 &/&  25.58 & 38.33 &/& 31.96 &/& 29.34  &/& 27.61 &/& 25.56 &\\

  &&& \textbf{NL-Ridge} \textit{(linear)}  &   \bf{38.19} &/& \bf{32.46} &/& \bf 30.00 &/& 28.41 &/& 26.73 &  \bf{37.67} &/& \bf{31.20} &/& 28.67 &/& 27.14 &/& 25.67 & \bf{38.56} &/& 32.53 &/& 29.90 &/& 28.13 &/& 26.29 &\\

 &&&  \textbf{NL-Ridge} \textit{(affine)}  &   38.17 &/& 32.42 &/& 29.98 &/& \bf{28.43} &/& \bf{26.79} &  37.65 &/& 31.18 &/& 28.68 &/& 27.16 &/& \bf{25.71} & 38.54 &/& \bf{32.54} &/& \bf{29.93} &/& \textbf{28.21} &/& \bf{26.40} &\\

  &&&  \textbf{NL-Ridge} \textit{(conical)}  &   38.03 &/& 32.16 &/& 29.72 &/& 28.07 &/& 26.49 & 37.46 &/& 30.86 &/& 28.38 &/& 26.85 &/& 25.43 &- &/& - &/& - &/& - &/& - &\\

 &&& \textbf{NL-Ridge} \textit{(convex)}  &   38.00 &/& 32.14 &/& 29.70 &/& 28.14 &/& 26.51 &  37.45 &/& 30.85 &/& 28.38 &/& 26.86 &/& 25.44 & -&/& - &/& - &/& - &/& - &\\[0.1cm]

\cdashline{3-32}\noalign{\vskip 0.1cm}

  &&\multirow{3}{*}{\begin{sideways} \scriptsize \textit{Multi-step} \end{sideways}} &   SS-GMM \cite{EPLL_unsupervised}  &  38.02 &/& 32.41 &/& 29.88 &/& 28.24 &/& 26.53 &  37.17 &/& 31.28 &/& 28.76 &/& 27.19 &/& 25.71 & - &/& - &/& - &/& - &/& - &\\
 

&&& TWSC \cite{TWSC}   &  38.17&/& 32.61  &/& 30.21 &/& 28.63 &/& 26.95 &  37.67 &/& 31.28 &/& 28.75 &/& 27.24 &/&  25.76 & - &/& - &/& -  &/& - &/& - &\\ 

 &&&  WNNM  \cite{WNNM} & \bf 38.36 &/&  \bf 32.70 &/& \bf 30.26 &/& \bf 28.69  &/&  \bf 27.05 &  \bf 37.80 &/& \bf 31.37  &/& \bf 28.83 &/& \bf 27.30 &/& \bf 25.87 & \bf 38.77 &/& \bf 32.97 &/& \bf 30.39 &/& \bf 28.70 &/& \bf 26.83 &\\



 \hline

\end{NiceTabular}%
}

\label{nlridge_resultsPSNR00}
\end{table*}

\begin{table*}[!t]
\centering
  \caption[The PSNR results of different methods on Darmstadt Noise Dataset (DND)]{The PSNR (dB) results on raw data on Darmstadt Noise Dataset (DND) \cite{DND}}
\resizebox{\columnwidth}{!}{%
  \begin{tabular}{c | cccccc | cccc} 
  \hline
 \multicolumn{1}{c}{\textit{}}  &  \multicolumn{6}{c}{\textit{Single-image}}  & \multicolumn{4}{c}{\textit{Dataset-based}} \\\hline
  \textbf{Methods} & BM3D \cite{BM3D} & NL-Bayes \cite{nlbayes} &  \textbf{NL-Ridge}  & KSVD \cite{ksvd} & NCSR  \cite{NCSR}  & WNNM \cite{WNNM}    & MLP \cite{mlp} & TNRD \cite{tnrd}  & FFDNet \cite{ffdnet} & DCT2net \cite{dct2net}    \\\hline
  \textbf{PSNR} (in dB) & \textbf{47.15}  &  46.84 &  47.12  & 46.87 & 47.07    & 47.05   & 45.71  & 45.70  & \textbf{47.40} & 46.83  \\\hline
\end{tabular}%
}
\label{nlrigde_dnd_res}
\end{table*}

\end{landscape}
\clearpage
}

\subsection{Results on test datasets}

We have evaluated the performance of NL-Ridge on artificially noisy images, as well as on real noisy images.

\subsubsection{Results on artificially noisy images corrupted by homoscedastic Gaussian noise}

We tested the denoising performance of our method on
three well-known datasets: Set12, BSD68 \cite{berkeley} and Urban100 \cite{urban}. A comparison with state-of-the-art algorithms is reported in  \Cref{nlridge_resultsPSNR00}. For the sake of a fair comparison, algorithms are divided into two categories: single-image methods, meaning that these methods (either traditional or deep learning-based) only have access to the input noisy image, and dataset-based methods (\textit{i.e.} supervised neural networks) that require a training phase beforehand and an external dataset. Note that only the single-image extension was considered for Noise2Self \cite{N2S} and the time-consuming ``internal adaptation'' option was not used for LIDIA \cite{LIDIA}. 

NL-Ridge, exclusively based on weighted aggregation of noisy patches, performs surprisingly at least as well as its traditional two-step counterparts \cite{BM3D, nlbayes}. It is particularly efficient on Urban100 dataset which contains abundant structural patterns and textures, achieving comparable performances with DnCNN \cite{dncnn} and FFDnet \cite{ffdnet},  popular supervised networks composed of hundreds of thousands of parameters. It is interesting to note that imposing constraints on the weights of combinations does not bring much improvement. It can even be detrimental in the case of conical or convex combinations of patches. This result is all the more surprising since many denoising algorithms are exclusively based on convex combinations of patches  \cite{nlmeans, PEWA, OWF}. We note however a slight superiority of the affine version over the unconstrained one at higher noise levels, which can be explained by our observation at the end of \Cref{internal_adaptation_nl_ridge_section}. In the rest, only the affine version of NL-Ridge is considered, which has the advantage of being normalization-equivariant in the case of Gaussian noise \cite{nenn}.

 \Cref{nlridge_photo} illustrates the visual results of different methods.  NL-Ridge  is very competitive with respect to well-established methods such as BM3D \cite{BM3D}. The self-similarity assumption is particularly useful to recover subtle details such as the stripes on the \textit{Barbara} image that are better reconstructed than DnCNN \cite{dncnn}.

\subsubsection{Results on real-world noisy images}

We tested the proposed method on the Darmstadt Noise Dataset \cite{DND} which is a dataset composed of $50$ real-noisy photographs. It relies on captures of a static scene with different ISO values, where the nearly noise-free low-ISO image is carefully post-processed to derive the ground-truth. In this challenge, the ground-truth images are not available. Each competitor submits the denoising results on the official website\footnote{https://noise.visinf.tu-darmstadt.de/}. The algorithms are then evaluated according to standard metrics and the ranking is made public\footnote{https://noise.visinf.tu-darmstadt.de/benchmark/}.


The real noise can be modeled as a Poisson-Gaussian noise:
\begin{equation}
    y \sim a \mathcal{P}(x / a) + \mathcal{N}(0, b)\,,
    \label{nlridge_poisson_gaussian_model}
\end{equation}
which can be further approximated with a heteroscedastic Gaussian noise whose variance is intensity-dependent:
\begin{equation}
    y \sim \mathcal{N}(x, \operatorname{diag}(ax+b))\,,
\label{nlridge_poisson_gaussian_model_approx}
\end{equation}
\noindent where $(a, b) \in \mathbb{R}^{+} \times \mathbb{R}^{+}$ are the noise parameters. For each noisy image, the authors \cite{DND} calculated the adequate noise parameters $(a, b)$ based on this model and made them available to the user. 
Note that for applying denoisers exclusively dedicated to homoscedastic Gaussian noise removal, a variance-stabilizing transformation (VST) such as the generalized Anscombe transform \cite{Anscombe, anscombe_kervrann} is performed beforehand. 
In our case,  stabilizing the variance is not necessary as NL-Ridge can handle mixed Poisson-Gaussian noise directly.

\begin{figure*}[t]
\centering
\renewcommand{\arraystretch}{0.5}
\begin{tabular}{cc}

\begin{tabular}{c}
\begin{tikzpicture}
\node[anchor=south west,inner sep=0] (image) at (0,0)
  {\includegraphics[width=0.252\textwidth]{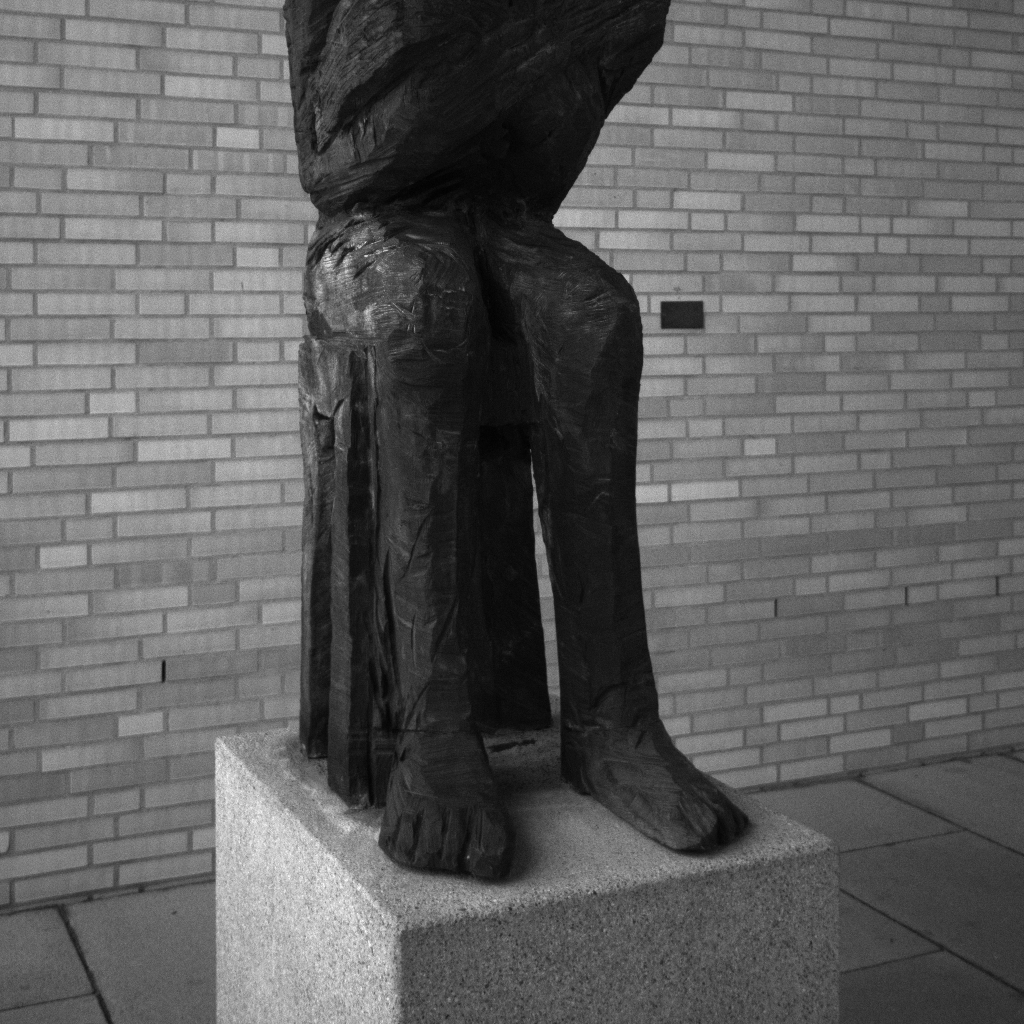}};
  

\node at (image.south east) [anchor=south east, inner sep=0] {\includegraphics[width=0.147\textwidth]{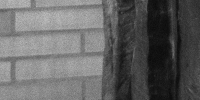}};

\begin{scope}[x={(image.south east)},y={(image.north west)}]
  \begin{scope}[shift={(0,1)},x={(1/1024,0)},y={(0,-1/1024)}]
  
   \draw[dodgerblue, semithick] (200, 450) rectangle (400, 550);

   \draw[dodgerblue, thick] (424, 724) rectangle (1024, 1024);

    \draw[densely dashed, thin, dodgerblue] (200, 550) -> (424, 1024);
    \draw[densely dashed, thin, dodgerblue] (400, 450) -> (1024, 724);
  \end{scope}
\end{scope}
\end{tikzpicture} \\
\scriptsize Noisy 
\end{tabular}
& 
\begin{tabular}{ccc}
\includegraphics[width=0.24\textwidth]{./Figures/original_crop0003.png} & \includegraphics[width=0.24\textwidth]{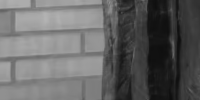} & \includegraphics[width=0.24\textwidth]{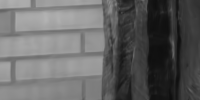}  \\
\scriptsize  Noisy & \scriptsize  BM3D \cite{BM3D} & \scriptsize  NL-Bayes \cite{nlbayes}  \\

\includegraphics[width=0.24\textwidth]{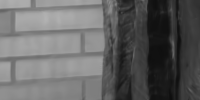} &
\includegraphics[width=0.24\textwidth]{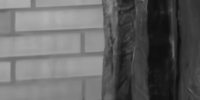}  & 
\includegraphics[width=0.24\textwidth]{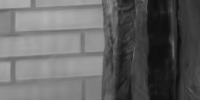} \\
 \scriptsize  WNNM \cite{WNNM} & \scriptsize FFDNet \cite{ffdnet}  & \scriptsize NL-Ridge (ours) \\
\end{tabular}

\end{tabular}
\caption[Qualitative comparison of image denoising results on real-world noisy images]{Qualitative comparison of image denoising results on real-world noisy images from Darmstadt Noise Dataset \cite{DND}. Zoom-in regions are indicated for each method.}
\label{nlridge_photo2}
\end{figure*}

 \Cref{nlridge_photo2} shows a qualitative comparison of the results obtained with state-of-the-art denoisers designed for the framework of additive white Gaussian noise. NL-Ridge does not suffer in comparison with more complex methods such as network-based ones \cite{ffdnet}. \Cref{nlrigde_dnd_res} compares the average PSNR values obtained on this dataset for different methods. NL-Ridge obtains comparable results with BM3D \cite{BM3D},  which is so far the best unsupervised method on this dataset.

\addtolength{\tabcolsep}{-5pt} 
\addtolength{\tabcolsep}{5pt}

\pgfdeclareplotmark{starBlue}{
    \node[star,star point ratio=2.25,minimum size=6pt,
          inner sep=0pt,draw=blue,solid,fill=blue] {};
}

\pgfdeclareplotmark{starRed}{
    \node[star,star point ratio=2.25,minimum size=6pt,
          inner sep=0pt,draw=red,solid,fill=red] {};
}

\pgfdeclareplotmark{starlichi}{
\node [star,star points=6, star point ratio=2.25, minimum size=6pt, inner sep=0pt,draw=color2,solid,fill=color2] {};
}


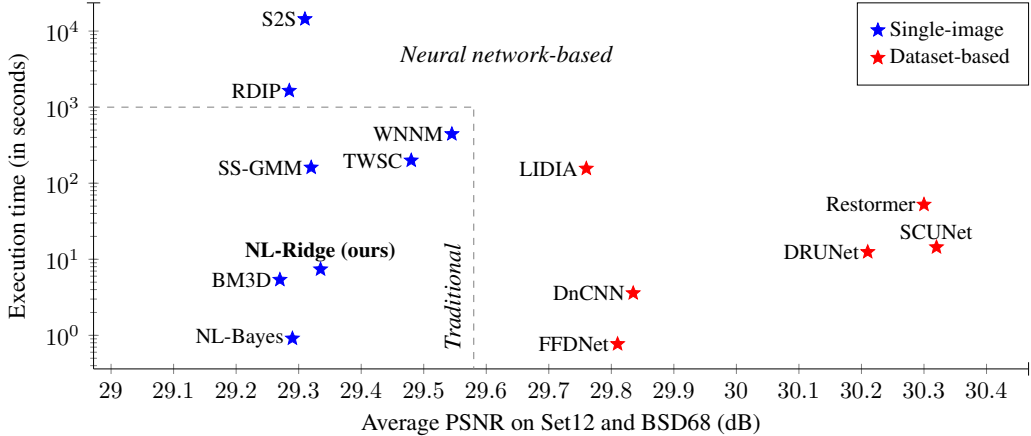
\begin{figure}[t]
\centering
\begin{tikzpicture}[scale=0.9]
\begin{axis}[
    title style={align=center},
    title={},
    cycle list name=exotic,
    ticks=both,
    ymode=log,
    ymin = 0.6,
    xmin = 29.04,
    xmax = 30.4,
    axis x line = bottom,
    axis y line = left,
    axis line style={-|},
    nodes near coords align={vertical},
    every node near coord/.append style={font=\tiny, xshift=-0.5mm},
    ylabel={Execution time (in seconds)},
    xlabel={Average PSNR on Set12 and BSD68 (dB)},
    legend style={at={(1, 1)}, anchor=north east, legend columns=1},
    every axis legend/.append style={nodes={right}, inner sep = 0.2cm},
    enlarge x limits=0.05,
    enlarge y limits=0.05,
    width=15.4cm,
    height=7cm,
]

    \addplot[only marks, blue,mark=starBlue, mark size=3pt ,mark options={solid}] coordinates {(29.335, 7.39)} node[midway, above] {\small \textbf{\textcolor{black}{NL-Ridge (ours)} }};

    \addplot[only marks, red,mark=starRed, mark size=3pt ,mark options={solid}] coordinates {(30.30, 52.39)} node[midway, left] {\small \textcolor{black}{Restormer}};

    
    \addplot[only marks, blue,mark=starBlue, mark size=3pt ,mark options={solid}] coordinates {(29.29, 0.911)} node[midway, left] {\small \textcolor{black}{NL-Bayes}};


    \addplot[only marks, blue,mark=starBlue, mark size=3pt ,mark options={solid}] coordinates {(29.27, 5.39)} node[midway, left] {\small \textcolor{black}{BM3D}};

    \addplot[only marks, blue,mark=starBlue, mark size=3pt ,mark options={solid}] coordinates {(29.545, 443.22)} node[midway, left] {\small \textcolor{black}{WNNM}};

    \addplot[only marks, red,mark=starRed, mark size=3pt ,mark options={solid}] coordinates {(30.21, 12.48)} node[midway, left] {\small \textcolor{black}{DRUNet}};



    \addplot[only marks, red,mark=starRed, mark size=3pt ,mark options={solid}] coordinates {(29.835, 3.59)} node[midway, left] {\small \textcolor{black}{DnCNN}};

    \addplot[only marks, blue,mark=starBlue, mark size=3pt ,mark options={solid}] coordinates {(29.31, 14400)} node[midway, left] {\small \textcolor{black}{S2S}};



    \addplot[only marks, blue,mark=starBlue, mark size=3pt ,mark options={solid}] coordinates {(29.285, 1645)} node[midway, left] {\small \textcolor{black}{RDIP}};

    \addplot[only marks, red,mark=starRed, mark size=3pt ,mark options={solid}] coordinates {(29.81, 0.77)} node[midway, left] {\small \textcolor{black}{FFDNet}};

    \addplot[only marks, red,mark=starRed, mark size=3pt ,mark options={solid}] coordinates {(29.76, 155.86)} node[midway, left] {\small \textcolor{black}{LIDIA}};

     \addplot[only marks, red,mark=starRed, mark size=3pt ,mark options={solid}] coordinates {(30.32, 14.43)} node[midway, above] {\small \textcolor{black}{SCUNet}};


    \addplot[only marks, blue,mark=starBlue, mark size=3pt ,mark options={solid}] coordinates {(29.32, 160.93)} node[midway, left] {\small \textcolor{black}{SS-GMM} };

    \addplot[only marks, blue,mark=starBlue, mark size=3pt ,mark options={solid}] coordinates {(29.48, 199.30)} node[midway, left] {\small \textcolor{black}{TWSC}};


    \addplot[dashed,color=gray]coordinates {(0,1000)(29.58,1000)};

    \addplot[dashed,
    color=gray]
    coordinates {
    (29.58,0.01)(29.58,1000)
    };

    \addplot[only marks,color=white,mark=*, mark size=3pt ,mark options={solid}] coordinates {(29.45, 5000)} node[midway, right] { \textcolor{black}{\textit{Neural network-based}}};

    \addplot[only marks,color=white,mark=*, mark size=3pt ,mark options={solid}] coordinates {(29.52, 3.2)} node[midway, right] {\begin{sideways} \textcolor{black}{\textit{Traditional}} \end{sideways}};

\legend{\small Single-image, \small Dataset-based}

\end{axis}
\end{tikzpicture}

\caption{The execution time on CPU for an image of size $512\times512$ v.s the average PSNR results on Set12 and BSD68 \cite{berkeley} for synthetic Gaussian noise with $\sigma=25$ of the most effective popular methods  \cite{drunet, dncnn, ffdnet, LIDIA, scunet, restormer, nlridge, nlbayes, BM3D, WNNM, EPLL_unsupervised, TWSC, rethinking, S2S}. These results are calculated based on \Cref{nlridge_resultsPSNR00}.}
\label{nlridge_complexity}
\end{figure}

\subsection{Complexity}

We want to emphasize that NL-Ridge, is relatively fast compared to its traditional and deep-learning-based single-image counterparts. The running times of different state-of-the-art algorithms are reported in \Cref{nlridge_complexity}. It is provided for information purposes only, as the implementation, the language used and the machine on which the code is run, highly influence the execution time. The CPU used is a 2,3 GHz Intel Core i7. Note that NL-Ridge has been entirely implemented in Python with Pytorch, enabling it to run on GPU as well unlike its traditional counterparts, making it even faster. 
It is worth noting that traditional single-image methods are much less computationally demanding than single-image deep-learning-based ones \cite{rethinking, S2S} that use time-consuming gradient descent algorithms for optimization, while traditional ones have closed-form solutions.

\section{Conclusion}

In this paper, we presented a unified view to reconcile two-step single-image non-local denoisers through the minimization of a risk from a family of estimators, exploiting unbiased risk estimates on the one hand and the ``internal adaptation'' on the other. We derive NL-Ridge algorithm, which leverages local linear combinations of noisy similar patches. Our experimental results show that NL-Ridge compares favorably with its state-of-the-art counterparts, including recent single-image deep network methods which are much more computationally demanding. Moreover, NL-Ridge is very versatile and can deal with a lot of different types of noise. To the best of our knowledge, NL-Ridge achieves state-of-the-art performance in the field of fast single-image denoising. Admittedly, more efficient methods in terms of PSNR values exist in the literature, notably WNNM \cite{WNNM}, but they are based on costly iterative schemes (involving a dozen steps), which may prove prohibitive in certain situations. An interesting line of research for future work is to study the benefits of iterating the linear combinations of patches while keeping a low computational burden. 

\newpage
\appendix

\section{Multivariate quadratic programming under affine constraints}
\label{appendix_useful_opti}

\begin{lemma}[Multivariate quadratic programming]
\label{lemma0}
Let $Q, C \in \mathbb{R}^{k \times k}$. If $Q$ is symmetric positive definite, 
$$
\left\{
    \begin{array}{l}
        \displaystyle \mathop{\arg \min}\limits_{\Theta \in \mathbb{R}^{k\times k}} \quad \operatorname{tr}\left(\frac{1}{2} \Theta^\top Q \Theta + C^\top \Theta \right) = - Q^{-1} C = I_k -  Q^{-1}  (Q+C)  
\,, \\
        \displaystyle \mathop{\arg \min}\limits_{
\substack{\Theta \in \mathbb{R}^{k\times k} \\ \text{s.t.    } \Theta^\top \mathbf{1}_k =  \mathbf{1}_k}} \operatorname{tr}\left(\frac{1}{2} \Theta^\top Q \Theta + C^\top \Theta \right)  =  I_k - \left( Q^{-1}  - \frac{Q^{-1} \mathbf{1}_k (Q^{-1} \mathbf{1}_k)^\top}{ \mathbf{1}_k^\top  Q^{-1}  \mathbf{1}_k} \right)(Q+C)   \,.

    \end{array}
\right.
$$

\end{lemma}

\begin{proof}
 Let $\theta_j$ and $c_j$ denote the $j^{th}$ column of matrix $\Theta \in \mathbb{R}^{k \times k}$ and $C \in \mathbb{R}^{k \times k}$, respectively.

\noindent First of all,
\begin{displaymath}\operatorname{tr}\left(\frac{1}{2} \Theta^\top Q \Theta + C^\top \Theta \right) = \sum_{j=1}^{k} \frac{1}{2} \theta_j^\top Q \theta_j + c_j^\top \theta_j = \sum_{j=1}^{k} h_j(\theta_j) \,.\end{displaymath}

\noindent with $h_j : \theta \in \mathbb{R}^{k} \mapsto \frac{1}{2} \theta^\top Q \theta + c_j^\top \theta$. The minimization problem is then separable and amounts to solve $k$ independent quadratic programming subproblems.
As $\operatorname{Hess} h_j(\theta) = Q$ which is symmetric positive definite, $h_j$ is strictly convex and so $h_j$ has at most one global minimum. By canceling the gradient, we have:
\begin{displaymath}\nabla h_j(\theta) = 0 \Leftrightarrow Q\theta  + c_j = 0 \Leftrightarrow \theta  = - Q^{-1} c_j\,.\end{displaymath}

\noindent Finally,
\begin{displaymath}\arg \min_{\Theta \in \mathbb{R}^{k \times k}} \operatorname{tr}\left(\frac{1}{2} \Theta^\top Q \Theta + C^\top \Theta \right)  =  -Q^{-1} C     
= I_k -  Q^{-1}  (Q+C) \,.\end{displaymath}

Moreover, according to the Karush–Kuhn–Tucker conditions, the minimizer $\theta^\ast$ of $h_j$ under the constraint $\mathbf{1}_k^\top \theta  = 1$ satisfies $\nabla h_j(\theta) = \lambda \mathbf{1}_k$
\noindent with $\lambda \in \mathbb{R}$. Thus, $Q\theta^\ast + c_j = \lambda \mathbf{1}_k$, hence, 
\begin{displaymath}\theta^\ast  = \lambda Q^{-1} \mathbf{1}_k - Q^{-1} c_j\,.\end{displaymath} 

\noindent Since $\mathbf{1}_k^\top \theta^\ast  = 1$, we deduce that 
$\mathbf{1}_k^\top \theta^\ast  = \lambda \mathbf{1}_k^\top  Q^{-1} \mathbf{1}_k - \mathbf{1}_k^\top Q^{-1} c_j  = 1\,$ then,
\begin{displaymath}\lambda = \frac{1 + \mathbf{1}_k^\top Q^{-1} c_j}{\mathbf{1}_k^\top  Q^{-1} \mathbf{1}_k}  = \frac{\mathbf{1}_k^\top  Q^{-1} (Q e_j + c_j)}{\mathbf{1}_k^\top  Q^{-1} \mathbf{1}_k} \,.\end{displaymath}
\noindent Finally, by noticing that $-Q^{-1} c_j = e_j - Q^{-1}(Q e_j + c_j) $
\begin{displaymath}
\theta^\ast  = \frac{\mathbf{1}_k^\top  Q^{-1} (Q e_j + c_j)}{\mathbf{1}_k^\top  Q^{-1} \mathbf{1}_k}  Q^{-1} \mathbf{1}_k - Q^{-1} c_j 
= e_j - \left(Q^{-1}-  \frac{Q^{-1} \mathbf{1}_k (Q^{-1} \mathbf{1}_k)^\top}{ \mathbf{1}_k^\top  Q^{-1}  \mathbf{1}_k} \right) (Qe_j +c_j) \,,\end{displaymath}

\noindent hence, 

\begin{displaymath}
\mathop{\arg \min}\limits_{\Theta \in \mathbb{R}^{k\times k} \text{s.t.    } \Theta^\top \mathbf{1}_k =  \mathbf{1}_k} \operatorname{tr}\left(\frac{1}{2} \Theta^\top Q \Theta + C^\top \Theta \right)  =  I_k - \left( Q^{-1}  - \frac{Q^{-1} \mathbf{1}_k (Q^{-1} \mathbf{1}_k)^\top}{ \mathbf{1}_k^\top  Q^{-1}  \mathbf{1}_k} \right)(Q+C)  \,.
\end{displaymath}

\end{proof}

\begin{lemma} 
Let $A \in \mathbb{R}^{n \times k}$ and $v \in \mathbb{R}^k \setminus \{ 0\}$. 
\begin{displaymath}\arg \min_{\beta \in \mathbb{R}^n} \|  A - \beta v^\top \|_F^2 = \frac{Av}{\| v \|_2^2} \end{displaymath}
\label{lemma_nlbayes}
\end{lemma}

\begin{proof} $ \|  A - \beta v^\top \|_F^2  = \sum_{i=1}^{n} \|  A_{i, \cdot} - \beta_i v \|_2^2 = \sum_{i=1}^{n} \|  A_{i, \cdot} \|_2^2  - 2 \beta_i \langle A_{i, \cdot}, v  \rangle + \beta_i^2 \|  v \|_2^2 \,. $
Now, as the univariate quadratic function $\beta_i \in \mathbb{R} \mapsto  \|  A_{i, \cdot} \|_2^2 - 2 \beta_i \langle A_{i, \cdot}, v  \rangle + \beta_i^2 \|  v \|_2^2 $ is minimized for $\beta_i = \langle A_{i, \cdot}, v  \rangle/ \| v \|_2^2$, we have $\displaystyle \arg \min_{\beta} \|  A - \beta v^\top \|_F^2 = Av / \| v \|_2^2$.
\end{proof}
\bigskip
\section{Mathematical proofs for NL-Ridge}
\label{appendix_nlridge}

In what follows, $X, Y \in \mathbb{R}^{n \times k}$. In each case, $Y_{i,j}$ follows a noise model which is centered on $X_{i,j}$  (\textit{i.e.} $\mathbb{E}(Y_{i,j}) = X_{i,j}$) and variables $Y_{i,j}$ are supposed independent along each row. More precisely, three types of noise are studied:
\begin{itemize}
    \item Gaussian noise: $Y_{i,j} \sim \mathcal{N}(X_{i,j}, V_{i,j})$ with $V \in (\mathbb{R}_\ast^+)^{n \times k}$ representing the noisemap, \textit{i.e.} the variance per pixel. In particular, for homoscedastic Gaussian noise,  $V = \sigma^2 \mathbf{1}_n \mathbf{1}_k^\top$ with $\sigma > 0$, that is $Y_{i,j} \sim \mathcal{N}(X_{i,j}, \sigma^2)$,  
    \item Poisson noise: $Y_{i,j} \sim \mathcal{P}(X_{i,j})$,
    \item Mixed Poisson-Gaussian noise: $Y_{i,j} \sim a \mathcal{P}(X_{i,j}/a) + \mathcal{N}(0, b)$ with $(a, b) \in (\mathbb{R}^+_\ast)^2$. 
\end{itemize}

\noindent The local denoiser in NL-Ridge is of the form $f_\Theta : Y \in \mathbb{R}^{n \times k}  \mapsto Y \Theta$  with $\Theta \in \mathbb{R}^{k \times k}$ and the quadratic risk is defined as $\mathcal{R}_\Theta(X) = \mathbb{E} \| f_\Theta(Y) - X  \|_F^2$.

\subsection{Minimization of the quadratic risk}

\begin{lemma}[A closed-form expression for the quadratic risk]
\label{lemma3}
Let $X, Y \in \mathbb{R}^{n \times k}$ and $V  \in (\mathbb{R^+_\ast})^{n \times k}$ such that the $Y_{i,j}$ are independent along each row,  
$\mathbb{E}(Y_{i,j}) = X_{i,j}$ and $\mathbb{V}(Y_{i,j}) = V_{i,j}$.

\begin{displaymath}\mathcal{R}_\Theta(X) = \mathbb{E} \| f_\Theta(Y) - X \|^2_F = \| X\Theta - X \|^2_F+  \operatorname{tr}\left(\Theta^\top \operatorname{diag}(V^\top \mathbf{1}_n) \Theta \right)\,.\end{displaymath}
\end{lemma}

\begin{proof}

By development of the squared Frobenius norm:
\begin{displaymath}\| Y \Theta - X \|_F^2  = \| Y \Theta \|_F^2 + \|  X \|_F^2 - 2 \langle Y \Theta ,  X  \rangle_F\,.\end{displaymath}

\noindent Now by linearity of expectation $\mathbb{E}\langle Y \Theta ,  X  \rangle_F = \langle X \Theta ,  X  \rangle_F\,,$ and, as $Y_{i,j}$ are independent along each row, and as $\mathbb{E}(Y_{i,j}^2) = \mathbb{E}(Y_{i,j})^2 + \mathbb{V}(Y_{i,j}) = X_{i,j}^2 + V_{i,j}$, we have:
\begingroup \allowdisplaybreaks \begin{align*}
\mathbb{E}\| Y \Theta \|_F^2 &= \mathbb{E} \left( \sum_{i=1}^{n} \sum_{j=1}^{k} \left( \sum_{l=1}^{k} Y_{i,l} \Theta_{l,j} \right)^2 \right) \\
&= \sum_{i=1}^{n} \sum_{j=1}^{k}  \mathbb{E}\left(\left( \sum_{l=1}^{k} Y_{i,l} \Theta_{l,j} \right)^2 \right) \\
&= \sum_{i=1}^{n} \sum_{j=1}^{k}  \left( \sum_{l=1}^{k} (X_{i,j}^2 + V_{i,j}) \Theta_{l,j}^2 + 2 \sum_{1 \leq l < l' \leq k} X_{i,l} \Theta_{l,j} X_{i,l'} \Theta_{l',j}  \right) \\
&= \sum_{i=1}^{n} \sum_{j=1}^{k}  \left( \sum_{l=1}^{k} V_{i,j} \Theta_{l,j}^2 + \sum_{\substack{1 \leq l \leq k \\ 1 \leq l' \leq k}} X_{i,l} \Theta_{l,j} X_{i,l'} \Theta_{l',j}  \right) \\
&= \sum_{i=1}^{n} \sum_{j=1}^{k} \sum_{l=1}^{k} V_{i,j} \Theta_{l,j}^2 + \sum_{i=1}^{n} \sum_{j=1}^{k} \left(\sum_{l=1}^{k} X_{i,l} \Theta_{l,j} \right)^2  \\
&=  \sum_{j=1}^{k} \sum_{l=1}^{k} \Theta_{l,j} \left(\sum_{i=1}^{n} V_{i,j} \right)\Theta_{l,j} + \| X \Theta \|_F^2  \\
&= \operatorname{tr}\left(\Theta^\top \operatorname{diag}(V^\top \mathbf{1}_n) \Theta \right)  + \| X \Theta \|_F^2 \,.
\end{align*}%
\endgroup

\noindent Hence, \begin{displaymath}\mathbb{E} \| Y \Theta - X \|_F^2  = \| X \Theta - X\|_F^2 + \operatorname{tr}\left(\Theta^\top \operatorname{diag}(V^\top \mathbf{1}_n) \Theta \right)\,.\end{displaymath}

\end{proof}

\begin{proposition}[Minimization of the quadratic risk]
\label{prop_nlridge_minimisation_risk}
Let $\mathcal{R}_\Theta(X) = \mathbb{E} \| f_\Theta(Y) - X  \|_F^2$ the quadratic risk and $Q = X^\top X + D$
with $D$ a diagonal matrix defined as:
\begin{displaymath}D = \left\{
    \begin{array}{ll}
        n\sigma^2 I_k & \small \mbox{(for homoscedastic Gaussian noise)} \\
        \operatorname{diag}(V^\top \mathbf{1}_n ) & \small \mbox{(for heteroscedastic Gaussian noise)} \\
        \operatorname{diag}(X^\top \mathbf{1}_n ) & \small \mbox{(for Poisson noise)} \\
        \operatorname{diag}((aX+b)^\top \mathbf{1}_n ) &\small  \mbox{(for mixed Poisson-Gaussian noise)} \\
        
    \end{array}
\right.\,.\end{displaymath}
If $Q$ is positive definite:
$$
\left\{
    \begin{array}{l}
        \displaystyle \mathop{\arg \min}\limits_{\Theta \in \mathbb{R}^{k\times k}} \quad  \mathcal{R}_\Theta(X)  = I_k - Q^{-1} D\,, \\
       \displaystyle \mathop{\arg \min}\limits_{
\substack{\Theta \in \mathbb{R}^{k\times k} \\ \text{s.t.    } \Theta^\top \mathbf{1}_k =  \mathbf{1}_k}} \mathcal{R}_\Theta(X)  = I_k - \left[Q^{-1} - \frac{Q^{-1} \mathbf{1}_k (Q^{-1} \mathbf{1}_k)^\top}{\mathbf{1}_k^\top  Q^{-1}  \mathbf{1}_k}   \right] D\,. 
    \end{array}
\right.$$

\end{proposition}

\begin{proof}
By  \Cref{lemma3},  
\begingroup \allowdisplaybreaks \begin{align*}
\mathcal{R}_\Theta(X) &= \| X\Theta - X \|^2_F  + \operatorname{tr}\left(\Theta^\top \operatorname{diag}(V^\top \mathbf{1}_n) \Theta\right) \\
&= \operatorname{tr}\left((X\Theta - X)^\top (X\Theta - X)\right) + \operatorname{tr}\left(\Theta^\top \operatorname{diag}(V^\top \mathbf{1}_n) \Theta\right)\\
&= \operatorname{tr}\left(\Theta^\top X^\top X \Theta - 2 X^\top X \Theta + X^\top X + \Theta^\top \operatorname{diag}(V^\top \mathbf{1}_n) \Theta\right)\\
&= \operatorname{tr}\left(\Theta^\top (X^\top X + \operatorname{diag}(V^\top \mathbf{1}_n)) \Theta - 2 X^\top X \Theta\right) + \operatorname{tr}\left(X^\top X\right)
\end{align*} \endgroup 
 \Cref{lemma0} allows to conclude.
\end{proof}

\subsection{Unbiased risk estimates (URE)}
\label{appendix_nlridge_URE}

The three following propositions introduce unbiased risk estimates for $\mathcal{R}_\Theta(X)$ depending on the noise model assumed on $Y$, denoted  $\operatorname{URE}_\Theta(Y)$ in a generic way.    

\begin{proposition}[Gaussian noise] An unbiased estimate of the risk $\mathcal{R}_\Theta(X) =  \mathbb{E}\| f_\Theta(Y)  - X  \|_F^2$
 is:
 \begin{displaymath} \operatorname{SURE}_\Theta(Y) = \| Y \Theta - Y  \|_F^2 + 2
 \operatorname{tr}(D \Theta) - \operatorname{tr}(D)\,, \end{displaymath}
 \noindent with $D = \operatorname{diag}(V^\top \mathbf{1}_n)$.
 In particular, for homoscedastic Gaussian noise, $\operatorname{SURE}_\Theta(Y) = \| Y \Theta - Y  \|_F^2 + 2 n\sigma^2 \operatorname{tr}(\Theta) - nk\sigma^2\,.$
   \label{proposition_gURE}
\end{proposition}

\begin{proof}
For $n=1$, all components of $Y$ are independent and generalized Stein's unbiased risk estimate (SURE) \cite{SURE} is given by:
\begin{displaymath} \operatorname{SURE}_{\Theta}(Y) = \| f_{\Theta}(Y) - Y \|_F^2 
+ 2 \operatorname{tr}(\operatorname{diag}(V^\top \mathbf{1}_n) \Theta) - \operatorname{tr}(\operatorname{diag}(V^\top \mathbf{1}_n))\,.\end{displaymath}

\noindent For $n \geq 1$,  
\begin{displaymath} 
\mathbb{E}\| f_{\Theta}(Y) - X \|_F^2    
   =  \sum_{i=1}^{n} \mathbb{E} \| Y_{i, \cdot} \Theta - X_{i, \cdot} \|_F^2   
 =  \sum_{i=1}^{n} \mathbb{E} \left(\operatorname{SURE}_{\Theta}(Y_{i, \cdot})\right) 
 =  \mathbb{E} \left(  \sum_{i=1}^{n} \operatorname{SURE}_{\Theta}(Y_{i, \cdot})\right)\,,\end{displaymath}
 hence, 
 \begin{displaymath} \begin{aligned}
     \operatorname{SURE}_{\Theta}(Y) = \sum_{i=1}^{n} \operatorname{SURE}_{\Theta}(Y_{i, \cdot}) &= \sum_{i=1}^{n}   \| Y_{i, \cdot} \Theta - Y_{i, \cdot} \|_F^2 + 2 \operatorname{tr}(\operatorname{diag}(V_{i, \cdot}^\top) \Theta) - \operatorname{tr}(\operatorname{diag}(V_{i, \cdot}^\top))\\
     &= \| Y \Theta - Y \|_F^2 + 2 \operatorname{tr}(D \Theta) - \operatorname{tr}(D)\,. \end{aligned}\end{displaymath}
\end{proof}

\begin{proposition}[Poisson noise] An unbiased estimate of the risk $\mathcal{R}_\Theta(X) =  \mathbb{E} \| f_\Theta(Y)  - X  \|_F^2$
 is:
 \begin{displaymath}\operatorname{PURE}_\Theta(Y) = \| Y \Theta - Y  \|_F^2 
 + 2
 \operatorname{tr}(D \Theta) - \operatorname{tr}(D)\,,\end{displaymath}
 \noindent with $D = \operatorname{diag}(Y^\top \mathbf{1}_n)$.
  \label{proposition_pURE}
\end{proposition}

\begin{proof}
For $n=1$, all components of $Y$ are independent and Poisson unbiased risk estimate (PURE) \cite{PURE, PGURE} is given by:
\begin{displaymath} \operatorname{PURE}_{\Theta}(Y) = \| f_{\Theta}(Y) \|_F^2 + \| Y \|_F^2 -  Y \mathbf{1}_k -  2 \langle f^{[-1]}_{\Theta}(Y), Y \rangle_F\end{displaymath}
\noindent with $f^{[-1]}_\Theta$ is such that $f^{[-1]i}_{\Theta}(Y) = f^{i}_{\Theta}(Y - e_i)$.
We have: \begingroup \allowdisplaybreaks \begin{align*}
    \langle f^{[-1]}_{\Theta}(Y), Y \rangle_F &= \sum_{j=1}^{k} \left(\sum_{l=1}^{k} (Y_{1, l} - \delta_{l,j}) \Theta_{l,j} \right) Y_{1,j} \\&= \sum_{j=1}^{k} \left( \sum_{l=1}^{k} Y_{1, l} \Theta_{l,j} \right) Y_{1,j}  - \sum_{j=1}^{k} \left( \sum_{l=1}^{k} \delta_{l,j} \Theta_{l,j}  \right) Y_{1,j}  \\
    &= \langle Y\Theta, Y \rangle_F -  \sum_{j=1}^{k} \Theta_{j,j} Y_{1,j}= \langle Y\Theta, Y \rangle_F -  Y\operatorname{diag}(\Theta)\,.\end{align*}%
    \endgroup

\noindent So finally, $\displaystyle \operatorname{PURE}_{\Theta}(Y) = \| Y\Theta - Y \|_F^2  -  Y \mathbf{1}_k +  2 Y\operatorname{diag}(\Theta) = \| Y\Theta - Y \|_F^2 + 2\operatorname{tr}(D \Theta) - \operatorname{tr}(D).$

\noindent For $n \geq 1$,  
\begin{displaymath}\displaystyle
\mathbb{E}\| f_{\Theta}(Y) - X \|_F^2    
   =  \sum_{i=1}^{n} \mathbb{E} \| Y_{i, \cdot} \Theta - X_{i, \cdot} \|_F^2   
 =  \sum_{i=1}^{n} \mathbb{E} ( \operatorname{PURE}_{\Theta}(Y_{i, \cdot})) 
 =  \mathbb{E} \left(  \sum_{i=1}^{n} \operatorname{PURE}_{\Theta}(Y_{i, \cdot})\right)\,,\end{displaymath}
 hence, 
 \begin{displaymath}\begin{aligned}
     \operatorname{PURE}_{\Theta}(Y) = \sum_{i=1}^{n} \operatorname{PURE}_{\Theta}(Y_{i, \cdot}) &= \sum_{i=1}^{n}   \| Y_{i, \cdot} \Theta - Y_{i, \cdot} \|_F^2 + 2 \operatorname{tr}(\operatorname{diag}(Y_{i, \cdot}^\top) \Theta) - \operatorname{tr}(\operatorname{diag}(Y_{i, \cdot}^\top))\\
     &= \| Y \Theta - Y \|_F^2 + 2 \operatorname{tr}(D \Theta) - \operatorname{tr}(D)\,. \end{aligned}\end{displaymath}
\end{proof}

\begin{proposition}[Mixed Poisson-Gaussian noise]
An unbiased estimate of the risk $\mathcal{R}_\Theta(X) =  \mathbb{E} \| f_\Theta(Y)  - X  \|_F^2$
 is:
 \begin{displaymath}\operatorname{PG-URE}_\Theta(Y) = \| Y\Theta - Y \|_F^2   
 + 2\operatorname{tr}(D \Theta) - \operatorname{tr}(D)\,,\end{displaymath}
 \noindent with $D = \operatorname{diag}((aY+b)^\top \mathbf{1}_n)$.
 \label{proposition_mpgURE}
\end{proposition}

\begin{proof}
For $n=1$, all components of $Y$ are independent and the Poisson-Gaussian unbiased risk estimate (PG-URE) \cite{PGURE} is given by:
\begin{displaymath}\operatorname{PG-URE}_{\Theta}(Y) = \| f_{\Theta}(Y) \|_F^2 + \| Y \|_F^2 -  2 \langle f^{[-a]}_{\Theta}(Y), Y \rangle_F  -  (a Y + b) \mathbf{1}_k + 2 b \operatorname{div}(f_\Theta^{[-a]})(Y)\end{displaymath}
\noindent with $f^{[-a]}_\Theta$ is such that $f^{[-a]i}_{\Theta}(Y) = f^{i}_{\Theta}(Y - a e_i)$.
\begin{displaymath}\begin{aligned}
    \langle f^{[-a]}_{\Theta}(Y), Y \rangle_F &= \sum_{j=1}^{k} \left(\sum_{l=1}^{k} (Y_{1, l} - a \delta_{l,j}) \Theta_{l,j} \right) Y_{1,j} \\&= \sum_{j=1}^{k} \left( \sum_{l=1}^{k} Y_{1, l} \Theta_{l,j} \right) Y_{1,j}  - \sum_{j=1}^{k} \left( \sum_{l=1}^{k} a \delta_{l,j} \Theta_{l,j}  \right) Y_{1,j}  \\
    &= \langle Y\Theta, Y \rangle_F -  a \sum_{j=1}^{k} \Theta_{j,j} Y_{1,j}= \langle Y\Theta, Y \rangle_F -  a Y\operatorname{diag}(\Theta)\,. \end{aligned}\end{displaymath}

\noindent and $ \displaystyle \operatorname{div}(f^{[-a]}_{\Theta})(Y) = \sum_{j=1}^{k} \frac{\partial f^{[-a]j}_\Theta}{\partial y_j}(Y) = \sum_{j=1}^{k} \frac{\partial}{\partial y_j} f^{j}_{\Theta}(Y -a e_j) = \sum_{j=1}^{k} \Theta_{j,j} = \mathbf{1}_k^\top \operatorname{diag}(\Theta)    \,.$

\noindent So finally, 
$\displaystyle \operatorname{PG-URE}_{\Theta}(Y) =\| Y\Theta - Y \|_F^2  -  (aY+b) \mathbf{1}_k +  2 (a Y+b)\operatorname{diag}(\Theta)=  \| Y\Theta - Y \|_F^2 + 2\operatorname{tr}(D \Theta) - \operatorname{tr}(D).$

\noindent For $n \geq 1$,  \begin{displaymath} \mathbb{E}\| f_{\Theta}(Y) - X \|_F^2    
   =  \sum_{i=1}^{n} \mathbb{E} \| Y_{i, \cdot} \Theta - X_{i, \cdot} \|_F^2   
 =  \sum_{i=1}^{n} \mathbb{E} ( \operatorname{PG-URE}_{\Theta}(Y_{i, \cdot})) 
=\mathbb{E} \left( \sum_{i=1}^{n}  \operatorname{PG-URE}_{\Theta}(Y_{i, \cdot})\right)\,,\end{displaymath}
\noindent hence,
\begin{displaymath}\begin{aligned}
     \operatorname{PG-URE}_{\Theta}(Y) &= \sum_{i=1}^{n} \operatorname{PG-URE}_{\Theta}(Y_{i, \cdot}) \\&= \sum_{i=1}^{n}   \| Y_{i, \cdot} \Theta - Y_{i, \cdot} \|_F^2 + 2 \operatorname{tr}(\operatorname{diag}((aY_{i, \cdot}+b)^\top) \Theta) - \operatorname{tr}(\operatorname{diag}((aY_{i, \cdot} + b)^\top))\\
     &= \| Y \Theta - Y \|_F^2 + 2 \operatorname{tr}(D \Theta) - \operatorname{tr}(D)\,. \end{aligned}\end{displaymath}
\end{proof}

In the following, we denote $\operatorname{URE}_\Theta(Y)$ either the $\operatorname{SURE}_\Theta(Y)$, $\operatorname{PURE}_\Theta(Y)$ or $\operatorname{PG-URE}_\Theta(Y)$ estimate, depending on the noise model assumed on $Y$.

\begin{proposition}[Minimization of the URE]
 \label{proposition_minURE}
Let $Q = Y^\top Y$ and $D$ a positive diagonal one defined as:
\begin{displaymath}D = \left\{
    \begin{array}{ll}
        n\sigma^2 I_k & \mbox{(for homoscedastic Gaussian noise)} \\
        \operatorname{diag}(V^\top \mathbf{1}_n ) & \mbox{(for heteroscedastic Gaussian noise)} \\
        \operatorname{diag}(Y^\top \mathbf{1}_n ) & \mbox{(for Poisson noise)} \\
        \operatorname{diag}((aY+b)^\top \mathbf{1}_n ) & \mbox{(for mixed Poisson-Gaussian noise)} \\
        
    \end{array}
\right.\,.\end{displaymath}

\noindent If $Q$ is definite positive,

$$
\left\{
    \begin{array}{l}
       \displaystyle \mathop{\arg \min}\limits_{\Theta \in \mathbb{R}^{k\times k}} \quad \operatorname{URE}_\Theta(Y)  = I_k - Q^{-1} D\,, \\
        \displaystyle \mathop{\arg \min}\limits_{
\substack{\Theta \in \mathbb{R}^{k\times k} \\ \text{s.t.    } \Theta^\top \mathbf{1}_k =  \mathbf{1}_k}} \operatorname{URE}_\Theta(Y)  = I_k - \left[Q^{-1} - \frac{Q^{-1} \mathbf{1}_k (Q^{-1} \mathbf{1}_k)^\top}{\langle Q^{-1}, \mathbf{1}_k \mathbf{1}_k^\top \rangle_F}   \right] D\,.
    \end{array}
\right.$$

\begin{proof}
Using \Cref{proposition_gURE}, \ref{proposition_pURE} and \ref{proposition_mpgURE}, 
\begin{displaymath}\begin{aligned}\operatorname{URE}_\Theta(Y) &= \| Y\Theta - Y \|_F^2 +  2 \operatorname{tr}(D \Theta) - \operatorname{tr}(D)\\
&= \operatorname{tr}\left(     \Theta^\top Y^\top Y \Theta + 2( D - Y^\top Y) \Theta  \right) + \operatorname{const}\\
\end{aligned}\end{displaymath}
Lemma \ref{lemma0} allows to conclude.
\end{proof}

\end{proposition}

\begin{proposition}[URE for a noisier risk and its minimization]
\label{appendix_noisier}
Let $\alpha > 0$ and $W \in \mathbb{R}^{n \times k}$ with $W_{i,j} \sim \mathcal{N}(0, 1)$ independent along each row. We define the noisier risk as $\mathcal{R}_{\Theta}^{\text{Nr}, \alpha}(X) = \mathbb{E} \| f_\Theta(Y + \alpha W)  - X  \|_F^2$. An unbiased estimate of the noisier risk $\mathcal{R}_{\Theta}^{\text{Nr}, \alpha}(X)$ is:
\begin{displaymath}\operatorname{URE}_\Theta^{\text{Nr}, \alpha}(Y) =  \operatorname{URE}_\Theta(Y) +  n \alpha^2  \| \Theta \|_F^2\,.\end{displaymath}
\noindent and its minimization yields:
$$
\left\{
    \begin{array}{l}
       \displaystyle \mathop{\arg \min}\limits_{\Theta \in \mathbb{R}^{k\times k}} \quad \operatorname{URE}_\Theta^{\text{Nr}, \alpha}(Y)  = I_k - Q^{-1} (D +  n \alpha^2 I_k)   
\,, \\
        \displaystyle \mathop{\arg \min}\limits_{
\substack{\Theta \in \mathbb{R}^{k\times k} \\ \text{s.t.    } \Theta^\top \mathbf{1}_k =  \mathbf{1}_k}} \operatorname{URE}_\Theta^{\text{Nr}, \alpha}(Y)  = I_k - \left[Q^{-1} - \frac{Q^{-1} \mathbf{1}_k (Q^{-1} \mathbf{1}_k)^\top}{\mathbf{1}_k^\top  Q^{-1}  \mathbf{1}_k}   \right] (D +  n \alpha^2 I_k)\,,
    \end{array}
\right.$$

\noindent with $Q = Y^\top Y + n \alpha^2  I_k$ a symmetric definite-positive matrix and $D$ a diagonal one defined as:
\begin{displaymath}D = \left\{
    \begin{array}{ll}
    n \sigma^2 I_k & \mbox{(for homoscedastic Gaussian noise)} \\
        \operatorname{diag}(V^\top \mathbf{1}_n )  & \mbox{(for heteroscedastic Gaussian noise)} \\
        \operatorname{diag}(Y^\top \mathbf{1}_n)  & \mbox{(for Poisson noise)} \\
        \operatorname{diag}((aY+b)^\top \mathbf{1}_n )  & \mbox{(for mixed Poisson-Gaussian noise)} \\
    \end{array}
\right.\,.\end{displaymath}

\begin{proof}
As $f_\Theta$ is a linear function and $Y$ and $W$ are independent:
\begingroup \allowdisplaybreaks \begin{align*}
    \mathcal{R}_{\Theta}^{\text{Nr}, \alpha}(X) &= \mathbb{E} \| f_\Theta(Y + \alpha W)  - X  \|_F^2 \\
    &= \mathbb{E} \left[ \| f_\Theta(Y)  - X  \|_F^2 + \alpha^2 \| f_\Theta(W) \|_F^2  + 2 \langle f_\Theta(Y)  - X , \alpha f_\Theta(W) \rangle_F \right]\\
    &= \mathcal{R}_\Theta(X) + \alpha^2 \mathbb{E} \| f_\Theta(W) \|_F^2\\
    &= \mathbb{E} \left[ \operatorname{URE}_\Theta(Y) \right] + \alpha^2 \mathbb{E} \| W\Theta \|_F^2
    \end{align*}%
\endgroup

\noindent with, as the $W_{i,j}$ are independent along each row: \begin{displaymath} \mathbb{E}\| W \Theta \|_F^2 
=  \sum_{i=1}^{n} \sum_{j=1}^{k} \mathbb{E}  \left( \left( \sum_{l=1}^{k} W_{i,l} \Theta_{l,j} \right)^2 \right)   = \sum_{i=1}^{n} \sum_{j=1}^{k}  \sum_{l=1}^{k}  \Theta^{2}_{l,j} = n  \| \Theta\|_F^2 \,.\end{displaymath}

\noindent Now, using  \Cref{proposition_gURE}, \ref{proposition_pURE} and \ref{proposition_mpgURE},

\begin{displaymath}\begin{aligned}\operatorname{URE}_\Theta(Y) &= \| Y\Theta - Y \|_F^2 +  2 \operatorname{tr}(D \Theta) - \operatorname{tr}(D) + n \alpha^2   \| \Theta\|_F^2\\
&= \operatorname{tr}\left(     \Theta^\top (Y^\top Y + n\alpha^2 I_k) \Theta + 2( D - Y^\top Y) \Theta  \right) + \operatorname{const}\\
\end{aligned}\end{displaymath}

Lemma \ref{lemma0} allows to conclude.
\end{proof}

\label{prop_minURE_Nr}
\end{proposition}

\bigskip
\section{Mathematical proofs for NL-Bayes}
\label{sectionNLBayes}

In what follows, $X, Y \in \mathbb{R}^{n \times k}$, with $Y_{i,j} \sim \mathcal{N}(X_{i,j}, \sigma^2)$  independent along each column. The local denoiser in NL-Bayes is of the form $f_{\Theta, \beta} : Y \in \mathbb{R}^{n \times k} \mapsto  \Theta Y + \beta \mathbf{1}_k^\top$ with $\Theta \in \mathbb{R}^{n \times n}$ and $\beta \in \mathbb{R}^{n}$. The quadratic risk is defined as $\mathcal{R}_{\Theta, \beta}(X) = \mathbb{E} \| f_{\Theta, \beta}(Y) - X  \|_F^2$. We denote by $\mu_Z \in \mathbb{R}^{k}$ and $C_Z \in \mathbb{R}^{n \times n}$ the empirical mean and covariance matrix of a group of patches $Z \in \mathbb{R}^{n \times k}$, that is 
\begin{displaymath}\mu_Z = \frac{1}{k} Z \mathbf{1}_k 
\quad  \text{and} \quad 
C_Z  = \frac{1}{k} (Z  - \mu_Z \mathbf{1}_k^\top )  (Z  - \mu_Z \mathbf{1}_k^\top )^\top   \,. \end{displaymath}

\subsection{Minimization of the quadratic risk}

\begin{lemma}[A closed-form expression for the quadratic risk]
\label{lemma3_nlbayes}

\begin{displaymath}\mathcal{R}_{\Theta, \beta}(X) = \mathbb{E} \| f_{\Theta,  \beta}(Y) - X \|_F^2 = \|   \Theta X   - X +  \beta \mathbf{1}_k^\top \|_F^2 + k\sigma^2 \| \Theta \|_F^2 \,.\end{displaymath}
\end{lemma}

\begin{proof}
By development of the Frobenius norm and using Lemma \ref{lemma3}:
\begingroup \allowdisplaybreaks \begin{align*}
\mathbb{E} \| f_{\Theta,  \beta}(Y) - X \|_F^2 &= \mathbb{E} \left[ \| \Theta Y - X  \|_F^2 + \| \beta \mathbf{1}_k^\top \|_F^2 + 2 \langle \Theta Y - X, \beta \mathbf{1}_k^\top \rangle_F  \right] \\
&= \mathbb{E}  \| Y^\top \Theta^\top  - X^\top  \|_F^2   + \| \beta \mathbf{1}_k^\top \|_F^2 + 2 \mathbb{E}  \langle \Theta Y - X, \beta \mathbf{1}_k^\top \rangle_F  \\
&=  \| X^\top \Theta^\top  - X^\top  \|_F^2 + k \sigma^2 \| \Theta^\top    \|_F^2   + \| \beta \mathbf{1}_k^\top \|_F^2 + 2  \langle \Theta X - X, \beta \mathbf{1}_k^\top \rangle_F   \\
&=  \| X \Theta  - X + \beta \mathbf{1}_k^\top  \|_F^2 + k \sigma^2 \| \Theta \|_F^2  \\
\end{align*}%
\endgroup
\end{proof}

\begin{proposition}[Minimization of the quadratic risk] 
\begin{displaymath}\arg \min_{\substack{\Theta \in \mathbb{R}^{n \times n}\\ \beta \in \mathbb{R}^n}} \mathcal{R}_{\Theta, \beta}(X)  = \hat{\Theta},  \hat{\beta}\end{displaymath}
\noindent with $\hat{\Theta} = C_{X}(C_{X} + \sigma^2 I_n)^{-1}$ and $\hat{\beta}  = (I_n - \hat{\Theta})\mu_{X}$.
\label{prop1_nlbayes}
\end{proposition}

\begin{proof}
According to  \Cref{lemma3_nlbayes}, $\mathcal{R}_{\Theta, \beta}(X) = \mathbb{E} \| f_{\Theta,  \beta}(Y) - X \|_F^2 = \|   \Theta X   - X +  \beta \mathbf{1}_k^\top \|_F^2 + k\sigma^2 \| \Theta \|_F^2$.
\noindent For $\Theta$ fixed and using  \Cref{lemma_nlbayes}, it is minimized for $\displaystyle \beta = -(\Theta X - X)\mathbf{1}_k / k = (I_n - \Theta) \mu_X$.

\noindent Injecting it in the expression of the risk:
\begin{displaymath} \|   (X - \mu_X \mathbf{1}_k^\top)^\top \Theta^\top - (X - \mu_X \mathbf{1}_k^\top)^\top \|_F^2 + k\sigma^2 \|   \Theta^\top  \|_F^2\end{displaymath}
\noindent This quantity is minimal, using  \Cref{lemma0}, for \begin{displaymath} \hat{\Theta}^\top = I_n -  k\sigma^2 ((X - \mu_X \mathbf{1}_k^\top) (X - \mu_X \mathbf{1}_k^\top)^\top + k\sigma^2 I_n)^{-1} = I_n - k\sigma^2 (kC_X + k\sigma^2 I_n)^{-1} \,,\end{displaymath}
\noindent \textit{i.e.} 
\begin{displaymath}  \hat{\Theta} = I_n - \sigma^2 (C_X + \sigma^2 I_n)^{-1} = C_X (C_X+\sigma^2 I_n)^{-1} \,.\end{displaymath}
\end{proof}

\subsection{Unbiased risk estimate (URE)}

\begin{proposition}[Gaussian noise] An unbiased estimate of the risk $\mathcal{R}_{\Theta, \beta}(X) =  \mathbb{E} \| f_{\Theta, \beta}(Y)  - X  \|_F^2$
 is:
 \begin{displaymath}\operatorname{SURE}_{\Theta, \beta}(Y) = \| \Theta Y  - Y  + \beta \mathbf{1}_k^\top \|_F^2  + 2 k \sigma^2 \operatorname{tr}(\Theta) - nk\sigma^2\,.\end{displaymath}
 \label{sure_nlbayes}
\end{proposition}

\begin{proof}
For $k=1$, all components of $Y$ are independent and Stein's unbiased risk estimate (SURE) \cite{SURE} is given by:
\begin{displaymath}\operatorname{SURE}_{\Theta, \beta}(Y) = - n\sigma^2 + \| f_{\Theta, \beta}(Y) - Y \|_F^2 + 2\sigma^2 \operatorname{div} f_{\Theta, \beta}(Y)\end{displaymath}
with $\displaystyle \operatorname{div} f_{\Theta, \beta}(Y) = \sum_{i=1}^{n} \frac{\partial f_{\Theta, \beta}^{i}}{\partial y_{i}}(Y) = \sum_{i=1}^{n} \frac{\partial }{\partial y_{i}} \sum_{l=1}^{n} \Theta_{i,l} Y_{l,1} =  \sum_{i=1}^{n} \Theta_{i,i}= \operatorname{tr}(\Theta)$.

\noindent For $k \geq 1$,
\begin{displaymath} \mathbb{E}\| f_{\Theta, \beta}(Y) - X \|_F^2 
= \sum_{j=1}^{m} \mathbb{E} \| \Theta Y_{\cdot, j} + \beta   - X_{\cdot, j} \|_F^2  
= \sum_{j=1}^{k} \mathbb{E} ( \operatorname{SURE}_{\Theta, \beta}(Y_{ \cdot, j}))  = \mathbb{E}  \sum_{j=1}^{k} \operatorname{SURE}_{\Theta, \beta}(Y_{ \cdot, j}),\end{displaymath}
\noindent hence,
\begin{displaymath}\operatorname{SURE}_{\Theta, \beta}(Y) = \sum_{j=1}^{k} \operatorname{SURE}_{\Theta, \beta}(Y_{ \cdot, j}) = \| \Theta Y  - Y  + \beta \mathbf{1}_k^\top \|_F^2  + 2 k \sigma^2 \operatorname{tr}(\Theta) - nk\sigma^2\,.\end{displaymath} 
\end{proof}

\begin{proposition}[Minimization of the URE] 
\begin{displaymath}\arg \min_{\Theta, \beta} \operatorname{SURE}_{\Theta, \beta}(Y)   = \hat{\Theta}, \hat{\beta}\end{displaymath}
\label{sure_nlbayes_min}
\end{proposition}
\noindent with $\hat{\Theta}, \hat{\beta} =(C_Y - \sigma^2 I_n) C_Y^{-1},  (I_n - \hat{\Theta}) \mu_Y$.

\begin{proof}
For $\Theta$ fixed and using  \Cref{lemma_nlbayes}, $\operatorname{SURE}_{\Theta, \beta}(Y)$ is minimal for $\displaystyle \beta = -(\Theta Y - Y ) \mathbf{1}_k / k = (I_n- \Theta) \mu_Y$. Injecting it in the expression of SURE:
\begin{displaymath} \|   (Y - \mu_Y \mathbf{1}_k^\top)^\top \Theta^\top - (Y - \mu_Y \mathbf{1}_k^\top)^\top \|_F^2 + 2k\sigma^2 \operatorname{tr}(\Theta)  - nk\sigma^2\,.\end{displaymath}
\noindent This quantity is minimal, using  \Cref{lemma0}, for \begin{displaymath} \hat{\Theta}^\top = I_n -  k\sigma^2 ((Y - \mu_Y \mathbf{1}_k^\top) (Y - \mu_Y \mathbf{1}_k^\top)^\top)^{-1} = I_n - \sigma^2 C_Y^{-1} \,,\end{displaymath}
\noindent \textit{i.e.} 
\begin{displaymath}  \hat{\Theta} = (C_Y - \sigma^2 I_n) C_Y^{-1} \,.\end{displaymath}
\end{proof}

\bigskip
\section{Mathematical proofs for BM3D}
\label{sectionBM3D}

In what follows, $X, Y \in \mathbb{R}^{n \times k}$, with $Y_{i,j} \sim \mathcal{N}(X_{i,j}, \sigma^2)$  independent. The local denoiser in BM3D is of the form $f_{\Theta} : Y \mapsto  P^{-1} (\Theta \odot (P Y Q)) Q^{-1}$ with $\Theta \in \mathbb{R}^{n \times k}$ and $P \in \mathbb{R}^{n \times n}$ and $Q \in \mathbb{R}^{k \times k}$ are two orthogonal matrices (\textit{i.e.} $PP^\top = I_n$ and $QQ^\top = I_k$). The quadratic risk is defined as $\mathcal{R}_{\Theta}(X) = \mathbb{E} \| f_{\Theta}(Y) - X  \|_F^2$.

\subsection{Minimization of the quadratic risk}

\begin{lemma}[A closed-form expression for the quadratic risk]
\label{lemma3_bm3d}

$$\mathcal{R}_{\Theta}(X) = \mathbb{E} \| f_{\Theta}(Y) - X \|_F^2 = \|  (\Theta - \mathbf{1}_n \mathbf{1}_k^\top) \odot PXQ \|_F^2 + \sigma^2 \| \Theta \|_F^2 \,. $$
\end{lemma}

\begin{proof}
Let $W = Y -X$. We have $W_{i,j} \sim \mathcal{N}(0, \sigma^2)$. As $P$ and $Q$ are orthogonal matrices, they preserve the $\ell_2$ norm:
\begin{multline*}\|  f_{\Theta}(Y) - X \|_F^2  =   \|  \Theta \odot (PYQ) - PXQ \|_F^2 
=   \|  \Theta \odot (PXQ) + \Theta \odot (PWQ) - PXQ \|_F^2 \\
=  \|  (\Theta - \mathbf{1}_n \mathbf{1}_k^\top) \odot PXQ \|_F^2 + \| \Theta \odot  (PWQ) \|_F^2  + 2 \langle (\Theta - \mathbf{1}_n \mathbf{1}_k^\top) \odot PXQ , \Theta \odot  (PWQ) \rangle_F \,.
\end{multline*} 

\noindent Now computing the expected value for each term yields:
$$\mathbb{E} \langle (\Theta - \mathbf{1}_n \mathbf{1}_k^\top) \odot  (P X Q), \Theta \odot  (P W Q) \rangle_F = 0 $$
\noindent and
$$
\begin{aligned}
\mathbb{E} \| \Theta \odot  (PWQ) \|_F^2 = \sum_{i=1}^{n} \sum_{j=1}^{k} \mathbb{E} [(\Theta_{i,j} (PWQ)_{i,j})^2] & = \sum_{i=1}^{n} \sum_{j=1}^{k}  \mathbb{V} [\Theta_{i,j} (PWQ)_{i,j}] 
\\
&= \sum_{i=1}^{n} \sum_{j=1}^{k} \Theta_{i,j}^2 \mathbb{V} [ (PWQ)_{i,j}] 
\end{aligned} 
$$
with, as $W_{i,j}$ are independent and $P$ and $Q$ are orthogonal, 
$$
\begin{aligned}
\mathbb{V} [ (PWQ)_{i,j}] = \mathbb{V} \left( \sum_{l=1}^{k} \left(\sum_{l'=1}^{n} P_{i, l'} W_{l', l}  \right) Q_{l, j}  \right) 
&=   \sum_{l=1}^{k}  Q_{l, j}^2 \mathbb{V} \left(\sum_{l'=1}^{n} P_{i, l'} W_{l', l}  \right)   
\\
&=   \sum_{l=1}^{k}  Q_{l, j}^2  \sum_{l'=1}^{n} P_{i, l'}^2 \mathbb{V} \left(W_{l', l}  \right)  =\sigma^2\,. \end{aligned}
$$

\noindent Finally, \; $ \displaystyle \mathbb{E} \|  f_{\Theta}(Y) - X \|_F^2 =  \|  (\Theta - \mathbf{1}_n \mathbf{1}_k^\top) \odot PXQ  \|_F^2 + \sigma^2 \| \Theta \|_F^2$.

\end{proof}

\begin{proposition}[Minimization of the quadratic risk] 
$$\arg \min_{\Theta \in \mathbb{R}^{n \times k}} \mathcal{R}_{\Theta}(X)  = \frac{(PXQ)^{\odot 2}}{\sigma^2 + (PXQ)^{\odot 2}}\,.$$
\end{proposition}

\begin{proof}
According to Lemma \ref{lemma3_bm3d}, $\mathcal{R}_{\Theta}(X) = \mathbb{E} \| f_{\Theta}(Y) - X \|_F^2 = \|  (\Theta - \mathbf{1}_n \mathbf{1}_k^\top) \odot PXQ \|_F^2 + \sigma^2 \| \Theta \|_F^2$.

\noindent Let $\alpha \in \mathbb{R}$. The minimum of $x \mapsto \alpha^2(x - 1)^2 + \sigma^2 x^2$ is obtained for $\displaystyle x = \frac{\alpha^2}{\sigma^2 + \alpha^2}$. Finally,
$$  \arg \min_{\Theta} \;  \|  (\Theta - \mathbf{1}_n \mathbf{1}_k^\top) \odot PXQ  \|_F^2 + \sigma^2 \| \Theta \|_F^2 = \frac{(PXQ)^{\odot 2}}{\sigma^2 + (PXQ)^{\odot 2}}.$$
\end{proof}

\subsection{Unbiased risk estimate (URE)}

\begin{proposition}[Gaussian noise] An unbiased estimate of the risk $\mathcal{R}_{\Theta}(X) =  \mathbb{E} \| f_{\Theta}(Y)  - X  \|_F^2$
 is:
 $$\operatorname{SURE}_{\Theta}(Y) = \|  (\Theta - \mathbf{1}_n \mathbf{1}_k^\top) \odot  PYQ \|_F^2  +2\sigma^2 \langle \Theta, \mathbf{1}_n \mathbf{1}_k^\top \rangle_F -nk\sigma^2\,.$$
 \label{sure_bm3d}
\end{proposition}

\begin{proof}

\noindent Let $W = Y-X$. By development of the squared Frobenius norm, $ \; \displaystyle \|  f_{\Theta}(Y) - Y \|_F^2 = \|  f_{\Theta}(Y) - X \|_F^2 + \| W  \|_F^2 - 2 \langle f_{\Theta}(Y) - X, W \rangle_F.$ As $P^{-1} = P^\top$ and $Q^{-1} = Q^\top$:
$$\begin{aligned}\langle f_{\Theta}(Y), W \rangle_F &= \langle P^{-1} (\Theta \odot (P Y Q)) Q^{-1}, W \rangle_F \\
&= \langle \Theta \odot (P Y Q), PWQ \rangle_F  \\
&= \langle \Theta \odot (P X Q), PWQ  \rangle_F + \langle \Theta \odot (P W Q), PWQ  \rangle_F\,. \end{aligned}$$

\noindent Now computing the expected value for each term yields:
$$ \mathbb{E} \langle  \Theta \odot (PXQ) , PWQ \rangle_F = 0, \quad \mathbb{E} \| W  \|_F^2 = nk\sigma^2, \quad \mathbb{E} \langle X, W \rangle_F = 0$$ and $$ \mathbb{E} \langle  \Theta \odot (PWQ)  , PWQ \rangle_F  = \sigma^2 \langle \mathbf{1}_n \mathbf{1}_k^\top, \Theta \rangle_F.$$

\noindent Indeed, as the $W_{i,j}$ are independent and $P$ and $Q$ are orthogonal matrices, and according to the proof of Lemma \ref{lemma3_bm3d}:
$$
\mathbb{E} [\Theta_{i,j} (PWQ)_{i,j}^2] = \Theta_{i,j} \mathbb{E} [ (PWQ)_{i,j}^2] = \Theta_{i,j} \left(\underbrace{\mathbb{E} [ (PWQ)_{i,j}]^2}_{=0} + \underbrace{\mathbb{V} [ (PWQ)_{i,j}]}_{=\sigma^2} \right)  = \sigma^2 \Theta_{i,j}\,. $$

\noindent Finally, we get $\;  \mathbb{E} \|  f_{\Theta}(Y) - X \|_F^2 = \mathbb{E} \left[  \|  f_{\Theta}(Y) - Y \|_F^2  +2\sigma^2 \langle \mathbf{1}_n \mathbf{1}_k^\top, \Theta \rangle_F -nk\sigma^2 \right]$ with $\|  f_{\Theta}(Y) - Y \|_F^2 = \|  (\Theta - \mathbf{1}_n \mathbf{1}_k^\top) \odot  PYQ \|_F^2$.
\end{proof}

\begin{proposition}[Minimization of the URE] 
$$\arg \min_{\Theta \in \mathbb{R}^{n \times k}} \operatorname{SURE}_{\Theta}(Y)   = \mathbf{1}_n \mathbf{1}_k^\top - \frac{\sigma^2}{(PYQ)^{\odot 2}}\,,$$
and 
$$\arg \min_{\Theta \in \{0, 1\}^{n \times k}} \operatorname{SURE}_{\Theta}(Y)   = \mathds{1}_{\mathbb{R} \setminus [-\sqrt{2} \sigma, \sqrt{2} \sigma]}(PYQ)\,.$$
\label{sure_bm3d_min}
\end{proposition}

\begin{proof} 
$$\|  (\Theta - \mathbf{1}_n \mathbf{1}_k^\top) \odot  PYQ \|_F^2  +2\sigma^2 \langle \Theta, \mathbf{1}_n \mathbf{1}_k^\top \rangle_F = \sum_{i=1}^{n} \sum_{j=1}^{n} ( (PYQ)_{i,j} (\Theta_{i,j} - 1))^2  + 2\sigma^2 \Theta_{i,j} $$

\noindent Let $\alpha \in \mathbb{R}^\ast$. The minimum of $x \in \mathbb{R} \mapsto (\alpha (x -1))^2 + 2\sigma^2 x$ is obtained for $\displaystyle x_{\text{min}, 1} = 1 - \frac{\sigma^2}{\alpha^2}$. Hence, $$\arg \min_{\Theta \in \mathbb{R}^{n \times k}} \operatorname{SURE}_{\Theta}(Y)   = \mathbf{1}_n \mathbf{1}_k^\top - \frac{\sigma^2}{(PYQ)^{\odot 2}}\,.$$

\noindent The minimum of $x \in \{0, 1\} \mapsto (\alpha (x - 1))^2 + 2\sigma^2 x$ is obtained for $\displaystyle x_{\text{min}, 2} = \mathds{1}_{\mathbb{R} \setminus [-\sqrt{2} \sigma, \sqrt{2} \sigma]}(\alpha)$. Hence, $$\arg \min_{\Theta \in \{0, 1\}^{n \times k}} \operatorname{SURE}_{\Theta}(Y)   = \mathds{1}_{\mathbb{R} \setminus [-\sqrt{2} \sigma, \sqrt{2} \sigma]}(PYQ)\,.$$
\end{proof}

\section*{Acknowledgments}
This work was supported by Bpifrance agency (funding) through the LiChIE contract. Computations  were performed on the Inria Rennes computing grid facilities partly funded by France-BioImaging infrastructure (French National Research Agency - ANR-10-INBS-04-07, “Investments for the future”).

We would like to thank R. Fraisse (Airbus) for fruitful  discussions. 

\bibliographystyle{plain} 
\bibliography{references}

\end{document}